%% file: acl_latex.tex
\documentclass[11pt]{article}

\usepackage[final]{acl}

\usepackage{hyperref}
\usepackage{url}
\usepackage{times}
\usepackage{latexsym}
\usepackage{enumitem}
\usepackage{amsfonts}
\usepackage{ulem}
\usepackage{hyperref}
\usepackage{adjustbox}
\usepackage{algorithm}
\usepackage{algpseudocode}

\usepackage[T1]{fontenc}
\usepackage{booktabs}      
\usepackage{multirow}      
\usepackage{graphicx}      
\usepackage{makecell}      
\usepackage[table]{xcolor} 
\usepackage{amsmath}       
\usepackage{tcolorbox}
\usepackage{caption}
\usepackage{asymptote}
\usepackage{booktabs}
\usepackage{amsmath}
\usepackage{wrapfig}
\usepackage{lipsum}
\usepackage{array}
\usepackage{tabularx}
\usepackage{siunitx}   
\usepackage{booktabs}  
\usepackage[dvipsnames]{xcolor} 
\definecolor{skyblue}{RGB}{210,235,255}

\tcbuselibrary{breakable}  
\tcbuselibrary{skins}      
\definecolor{PineGreen}{HTML}{007C4A} 
\definecolor{OrangeRed}{HTML}{FF4500} 
\definecolor{mycolor}{gray}{0.92}    
\newcommand{\mymethod}{CircuitSeer}
\usepackage[utf8]{inputenc}

\usepackage{microtype}

\usepackage{inconsolata}
\usepackage{marvosym}
\usepackage{graphicx}
\definecolor{authorcolor}{RGB}{61, 134, 74}
\title{CircuitSeer: Mining High-Quality Data by Probing Mathematical Reasoning Circuits in LLMs}

\author{
  {\bf
    Shaobo Wang
    $^{\text{\Letter}}$$^*$$^{{\color{authorcolor}\boldsymbol{1,2}}}$
    Yongliang Miao
    $^*$$^{{\color{authorcolor}\boldsymbol{1}}}$
    Yuancheng Liu $^{{\color{authorcolor}\boldsymbol{1}}}$
  } \\
  {\bf    
    Qianli Ma
    $^{{\color{authorcolor}\boldsymbol{2}}}$    
    Ning Liao
    $^{{\color{authorcolor}\boldsymbol{2}}}$
    Linfeng Zhang
    $^{\text{\Letter}}$$^{{\color{authorcolor}\boldsymbol{1,2}}}$
  } \\
    {
    $^{\color{authorcolor}\boldsymbol{1}}$ EPIC Lab, SJTU $\quad$
    $^{\color{authorcolor}\boldsymbol{2}}$ Shanghai Jiao Tong University $\quad$ 
    } \\
    {
    * Equal contribution $\quad$ 
    \Letter\ Corresponding authors  $\quad$
    }
}

\begin{document}
\maketitle

\input{chapters/0_abstract}

\input{chapters/1_intro}

\input{chapters/2_related_work}

\input{chapters/3_method}

\input{chapters/4_exp}

\input{chapters/5_conclusion}

\clearpage
{
\small
\bibliography{custom}
}

\clearpage
\appendix
\section{Appendix}
\input{chapters/X_appendix}

\end{document}

%% file: chapters/0_abstract.tex
\begin{abstract}
Large language models (LLMs) have demonstrated impressive reasoning capabilities, but scaling their performance often relies on massive reasoning datasets that are computationally expensive to train on. Existing data selection methods aim to curate smaller, high-quality subsets but often rely on costly external models or opaque heuristics. In this work, we shift the focus from external heuristics to the model's internal mechanisms. We find that complex reasoning tasks consistently activate a sparse, specialized subset of attention heads, forming core reasoning circuits. Building on this insight, we propose CircuitSeer, a novel data selection method that quantifies the reasoning complexity of data by measuring its influence on these crucial circuits. Extensive experiments on 4 models and 9 datasets demonstrate CircuitSeer's superiority. Notably, fine-tuning Qwen2.5-Math-7B on just 10\% of data selected by our method achieves a 1.4-point gain in average Pass@1 over training on the full dataset, highlighting its efficiency and effectiveness.

\end{abstract}

%% file: chapters/1_intro.tex
\section{Introduction}
\label{sec:intro}
Large language models (LLMs), like OpenAI-o1~\citep{openai2024reason} and DeepSeek-R1~\citep{deepseekai2025deepseekr1incentivizingreasoningcapability}, have achieved remarkable advances in complex reasoning tasks, significantly contributing to the development of massive reasoning datasets. While datasets like OpenR1~\citep{openr1}, Synthetic-1~\citep{2025synthetic1}, and AM~\citep{zhao202514millionopensourcedistilled} provide essential resources for supervised fine-tuning (SFT) to further enhance LLM reasoning capabilities, their massive scale often demands substantial training time and computational resources. The problem of curating a more compact yet higher-quality reasoning dataset warrants further investigation.

\begin{figure}[tb!]
    \centering
    \includegraphics[width=\linewidth]{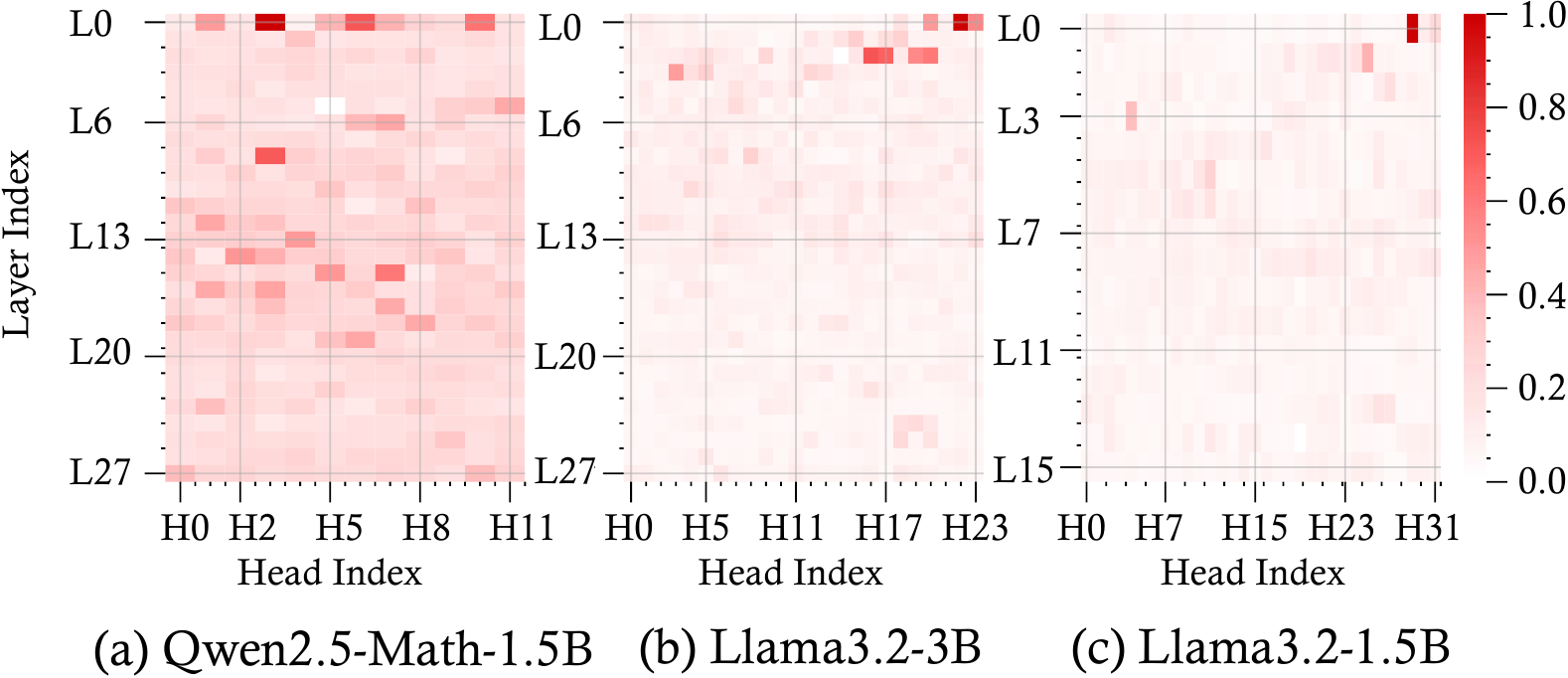}
    \caption{Heatmaps of attention-head importance during on probe reasoning dataset across four models: (a) Qwen2.5-Math-1.5B, (b) Llama3.2-3B, (c) Llama3.2-1B-Instruct. Brighter cells indicate higher contribution to reasoning; each model exhibits a sparse set of specialized reasoning heads, supporting our finding that only a subset of heads is for reasoning.}
    \label{fig:head_heatmaps}
    \vspace{-10pt}
\end{figure}

Existing methods, such as LIMO~\citep{ye2025limoreasoning}, s1~\citep{muennighoff2025s1simpletesttimescaling}, and COT-SELF-INSTRUCT~\citep{kim2023cot}, have attempted to curate high-quality subsets from large-scale data. However, these approaches suffer from several critical limitations. They often rely on complex and resource-intensive pipelines, such as rejection sampling that requires a separate, more powerful model, or model-based quality assessments that are computationally expensive. Furthermore, their core logic is frequently built upon intricate data engineering and qualitative heuristics for quality, difficulty, and diversity. This reliance on external models and heuristic rules makes these methods not only costly but also opaque, difficult to reproduce, and lacking a principled, quantitative basis for determining data quality.

\begin{figure*}[!t]
{
    \centering
    \vspace{-10pt}
\resizebox{.99\textwidth}{!}{
    \includegraphics[width=0.99\linewidth]{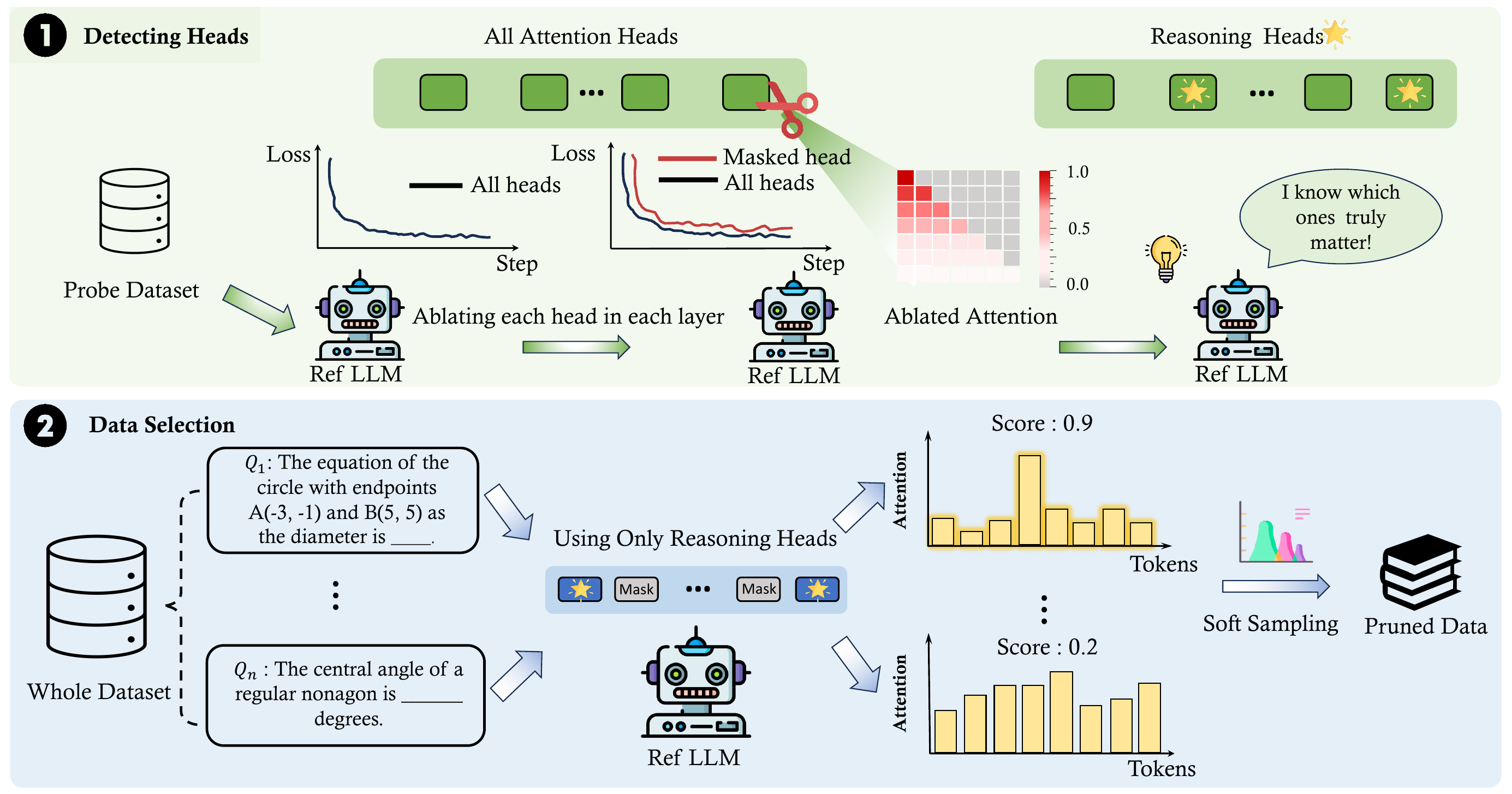}
}
\caption{
Overview of the CircuitSeer pipeline for reasoning-aware data selection.
\textbf{Stage~1: Detecting Heads.} The reference LLM is evaluated on a probe set while ablating each attention head by scaling its attention weights; heads whose ablation yield a significant increase in loss are retained as reasoning heads.
\textbf{Stage~2: Data Selection.} For each problem, we compute token-level attention distributions using only the detected heads and score samples with the formulation in Section~\ref{sec:scores} (Eq.~\eqref{score}. Scores parameterize soft sampling to form a compact, diverse subset enriched with reasoning-heavy examples.
}
\label{fig:pipeline}
\vspace{3pt}

\label{fig:cal_score}

}
\end{figure*}

To overcome these limitations, we shift our focus from external heuristics to the model's internal mechanisms, seeking a more fundamental signal for data quality. Through extensive experimentation, we observed a striking and consistent phenomenon: during complex reasoning tasks, only a small, specific subset of attention head circuits remain highly active. Crucially, we find this pattern of sparse, specialized activation to be a consistent trend across the diverse large lanuage models. This empirical finding aligns with the principles of Transformer circuits theory~\citep{anthropic_attribution_graphs, elhage2021framework, ameisen2025circuit}, which posits that different circuits emerge within transformer models to handle different tasks. Figure~\ref{fig:head_heatmaps} visually corroborates our observation, illustrating the sparse set of specialized attention heads that consistently activate during long reasoning, highlighting their pivotal role in this task.

Building on this key insight, we propose CircuitSeer, a novel method that quantifies the complex reasoning attribute of data by harnessing these direct, model-internal signals. Unlike traditional data selection strategies that depend on external evaluations or coarse heuristics, CircuitSeer offers a principled and quantifiable approach rooted in the model's own processing. Our pipeline for identifying these specialized reasoning heads and data mining is illustrated in Figure~\ref{fig:cal_score}. Our testing across multiple benchmarks and diverse models has demonstrated that CircuitSeer enhanced performance by selecting high quality data as shown in Figure \ref{fig:performance_comparison}.

To validate our approach, we conducted comprehensive ablation studies exploring the critical role of these reasoning heads. These experiments confirmed that scores derived from the dominant reasoning heads substantially improve model performance over using all heads indiscriminately, underscoring their functional specialization. Additionally, our used soft sampling proved more effective than simple top-k selection in enhancing both data diversity and model generalization. Our contributions can be summarized as follows:
\begin{enumerate}[leftmargin=10pt, topsep=-2pt, itemsep=1pt, partopsep=1pt, parsep=1pt]
    \item We identify a consistently emerging subset of reasoning-critical internal circuits (particularly attention heads) that underpin long-form mathematical reasoning, and we are the first to systematically harness these circuits to inform large-scale data selection.
    \item We utilize these crucial internal circuits to calculate a model-internal score that reflects the reasoning attribute of data , enabling a data selection method that achieves state-of-the-art results across multiple models.
    \item Through comprehensive experiments on 4 models and 9 test datasets, our method outperforms all baselines. Notably, on Qwen2.5-Math-7B, fine-tuning on just 10\% of the data selected by CircuitSeer surpassed training on the full dataset by 1.45 points in average Pass@1 and 5.54 points in average Majority@16.
\end{enumerate}

%% file: chapters/2_related_work.tex
\section{Related Works}
\label{sec:related}

\noindent\textbf{Data Selection.}
Data selection plays a crucial role in supervised fine-tuning (SFT) of the large language model(LLM), with recent efforts mainly focusing on either task relevance---prioritizing data similar to the target distribution through clustering, in-context signals, or reasoning filters \citep{shen2024rethinking,yang2024smalltolarge,xia2024rethinking,wang2025datawhisperer}---or importance-based selection, which estimates data contribution via loss, gradients, or optimization-driven scoring \citep{2020t5,zheng2023coverage,yang2024clip,xia2024rethinking,yang2024smalltolarge,shen2024rethinking}. However, fine-grained needs such as mathematical reasoning remain largely unexplored.

\noindent\textbf{LLM Reasoning.}
Recent work on reasoning in LLMs has emphasized test-time scaling~\citep{wei2022chain,lightman2023verify,snell2024scaling,welleck2024decoding}, showing that increasing the number of generated tokens or verification steps can substantially improve reasoning accuracy.
In parallel, open-source efforts have advanced the field by curating synthetic datasets with long reasoning trajectories~\citep{wang2023selfdiscover,sun2024logiclm} and distilling the reasoning skills of models such as DeepSeek-R1~\citep{deepseekai2025deepseekr1incentivizingreasoningcapability} into more compact LLMs~\citep{shridhar-etal-2023-distilling}.
Despite these advances, when dealing with large-scale reasoning corpora, the community still largely lacks efficient and principled, inherently interpretable data selection methods.

\noindent\textbf{Circuits Theory.} Mechanistic interpretability aims to uncover human-understandable algorithms within neural networks~\citep{olah2020zoom,vig-belinkov-2019-analyzing}. 
Within this paradigm, Circuits Theory~\citep{elhage2021framework,anthropic_attribution_graphs,anthropic_attribution_graphs_biology,lindner2023tracr} maps fine-grained connectivity patterns among attention heads and MLP neurons to algorithmic functions such as induction, composition, and copying. 
Recent extensions, including attribution graphs~\citep{anthropic_attribution_graphs,ameisen2025circuit}, trace causal information flows through these circuits, revealing components selectively activated by proxy tasks~\citep{mcgrath2023mechanistic}. CircuitSeer pinpoints reasoning-related circuits and further  exploits them to guide targeted data selection.

%% file: chapters/3_method.tex

\section{Methodology}

\subsection{Preliminaries}

Let $\mathcal{D} = \{(x_i, y_i)\}_{i=1}^N$ represent a comprehensive dataset designed for mathematical reasoning, where $N$ denotes the total number of samples. Each instance in this dataset is composed of an input problem $x_i$ and a corresponding solution process $y_i$, which includes both the detailed reasoning steps and the final answer. The objective is to identify an optimal subset $\mathcal{D}^{'} \subset \mathcal{D}$ that is suited for SFT.

Given a pretrained language model $\mathcal{M}$, fine-tuning is performed on the selected subset $\mathcal{D}^{'}$ to obtain a task-specialized model. The goal is to select $\mathcal{D}^{'}$ such that $\mathcal{M}$ achieves highest accuracy on an unseen test set $\mathcal{D}_{\text{test}}$, and competitive performance on standard mathematical reasoning benchmarks, comparable to fine-tuning on the full dataset. Formally, the selection problem is:
\begin{equation}
\mathcal{D}^{'} = \underset{\mathcal{D}^{'} \subseteq \mathcal{D}}{\arg\max}\,
\mathbb{E}_{(x, y) \sim \mathcal{D}_{\text{test}}}\,
\phi(\mathcal{M}(x), y)
\label{eq:objective}
\end{equation}
where $y$ denotes the ground-truth answer and
$\mathcal{\phi}(\cdot)$ represents a evaluation metric.

\subsection{Detecting Reasoning Circuits}
\label{sec:heads}
Circuit Theory~\citep{elhage2021framework,anthropic_attribution_graphs} shows that neurons and attention heads in Transformer models play distinct roles. Yet in large language models (LLMs), how to systematically identify and understand the attention circuits responsible for reasoning remains largely unexplored. Our initial analysis involves an ablation experiments of attention heads in LLMs, evaluated on a mathematical reasoning probe dataset.

We denote $\mathcal{M}_{\text{ref}}$ as the reference model used for attention head circuits selection and data scoring, $\mathcal{D}_{\text{probe}}$ as a curated probe set for evaluating circuit importance, and $\mathcal{H}$ be the set of all attention heads in $\mathcal{M}_{\text{ref}}$. For each circuit operationalized through a specific attention head $\tilde{h}$, we perform ablation to assess its contribution by employing undifferentiated attention \cite{zhou2024role}, effectively nullifying its influence. This is achieved by scaling the attention matrices—where $W_q$, $W_k$, and $W_v$ represent the query, key, and value matrices respectively, and $d_k$ denotes the dimension of the key vectors. Let $\mathcal{H}$ denote the set of all attention heads; we manipulate the attention weights to collapse into a lower-triangular matrix $A$. The matrix elements $a_{ij}$ are scaled by a small coefficient $\epsilon$:
\begin{equation}
\tilde{h}
  = \operatorname{Softmax}\!\left(
      \frac{\epsilon W_q W_k^{\top}}{\sqrt{d_k/|\mathcal{H}|}}
    \right) W_v
   \overset{\text{def}}{=} A W_v
\label{equation:undiff_attention}
\end{equation}

\begin{equation*}
\text{where } A = \bigl[a_{ij}\bigr],\quad
a_{ij} =
\begin{cases}
  \dfrac{1}{i}, & i \ge j,\\[4pt]
  0,            & i < j.
\end{cases}
\end{equation*}
The contribution of each candidate attention head is assessed by computing the expected increase in loss on the probe dataset induced by the intervention:
\begin{equation}
\mathbb{E}_{(x, y) \sim \mathcal{D}_{\text{probe}}} 
\left[\,\mathcal{L}(x, y;\mathcal{M}^{-c}_{\mathrm{ref}})
       - \mathcal{L}(x, y;\mathcal{M}_{\mathrm{ref}})\,\right],
\end{equation}
where $\mathcal{M}_{\mathrm{ref}}(x, y)$ denotes the loss of the reference model, and $\mathcal{M}^{-c}_{\mathrm{ref}}(x, y)$ denotes the loss after masking circuit $c$. 
These values are then sorted in descending order, and the top-$k$ heads are selected as the reasoning head set $\mathcal{H}_{\mathrm{math}}$.

\subsection{Using Reasoning Circuits To Select Data}
\label{sec:scores}

Building upon the reasoning circuits identified in Section~\ref{sec:heads}, we leverage them as data scorers to quantitatively evaluate the reasoning quality of each training sample $(x_i, y_i) \in \mathcal{D}$. For every sample, only the problem $x_i$ is input to the reference model  $\mathcal{M}_{\text{ref}}$, and scoring is performed exclusively using the selected reasoning circuits.

Let $x_i = (x_{i}^{(1)}, x_{i}^{(2)}, \ldots, x_{i}^{(n)})$ denote the input sequence of $n$ tokens. For each reasoning head $h \in \mathcal{H}_{\text{math}}$, the corresponding self-attention matrix $A^{h} \in \mathbb{R}^{n \times n}$ is extracted on model $\mathcal{M}_{ref}$, where $A^{h}_{j,k}$ represents the attention weight assigned from source token $j$ to target token $k$. The mean attention on token $k$ across all selected heads is:
\begin{equation}
    {\alpha}_k = \frac{1}{|\mathcal{H}_{\text{math}}|} \sum_{h \in \mathcal{H}_{\text{math}}}\sum_{j=1}^{n} A^{h}_{k,j},
\end{equation}
where $A^h$ is row-normalized such that $\sum_{j=1}^{n} A^{h}_{k,j} = 1$.

The score for the input $x_i$, denoted as $S(x_i)$, is the variance of the mean attention across all tokens:
\begin{equation}
    S(x_i) = \frac{1}{n} \sum_{k=1}^{n} \left({\alpha}_k -  \frac{1}{n} \sum_{k=1}^{n} {\alpha}_k \right)^2
\label{score}
\end{equation}

The score $S(x_i)$ is designed to measure the reasoning quality inherent in a sample. 
For problems that require multi-step reasoning, there often exist several logical key points 
where the model must form highly concentrated attention patterns and carry out step-by-step 
deductions around them. In contrast, for simple or formulaic problems, the attention distribution 
is more likely to be relatively uniform. A larger score $S(x_i)$ therefore indicates a higher 
level of reasoning quality of the sample.

Once all samples have been assigned scores, these scores are normalized to induce a probability distribution over the dataset. Specifically, each sample $x_i$ is assigned a selection probability proportional to its score $\frac{S(x_i)}{\sum_{(x_j,y_j) \in \mathcal{D}} S(x_j)}$.  We perform soft sampling over $\mathcal{D}$  according to this probability distribution until the desired size $|\mathcal{D}^{'}|$ is reached.

\input{tables/main_exp}

%% file: tables/main_exp.tex
\begin{table*}[tb!]
\centering
\vspace{-5pt}
\caption{Main evaluation results across our full suite of mathematical reasoning benchmarks. We compare various data selection methods by fine-tuning three base models. For AIME and AMC, we report both Pass@1 (P@1) and Majority@16 (M@16) accuracy (\%). For all other datasets, we report P@1. \textcolor{PineGreen}{$\uparrow$} indicates an improvement over the Random Sampling baseline, while \textcolor{OrangeRed}{$\downarrow$} indicates a degradation. The best performance among selection methods for each setting is in \textbf{bold}, and the second best is \uline{underlined}.}
\label{tab:main_results_detailed}
\vspace{-8pt}
\resizebox{\textwidth}{!}{
\begin{tabular}{l|cc|cc|cc|c|c|c|c|c|c|c}
\toprule
\multirow{2}{*}{\textbf{Method}} & \multicolumn{2}{c|}{\textbf{AIME24}} & \multicolumn{2}{c|}{\textbf{AIME25}} & \multicolumn{2}{c|}{\textbf{AMC23}} & \textbf{MATH} & \textbf{Olympiad} & \textbf{Kaoyan} & \textbf{GK 23} & \textbf{GK-Math} & \textbf{GK 24} & \textbf{Avg.} \\
\cmidrule(lr){2-3} \cmidrule(lr){4-5} \cmidrule(lr){6-7} \cmidrule(lr){8-8} \cmidrule(lr){9-9} \cmidrule(lr){10-10} \cmidrule(lr){11-11} \cmidrule(lr){12-12} \cmidrule(lr){13-13} \cmidrule(lr){14-14}
 & P@1 & M@16 & P@1 & M@16 & P@1 & M@16 & P@1 & P@1 & P@1 & P@1 & P@1 & P@1 & P@1 \\ \midrule
\multicolumn{14}{l}{\textbf{Qwen2.5-Math-7B-Base}} \\
\quad \textit{Zero-shot} & 1.0 & 0.0 & 0.4 & 3.3 & 13.4 & 27.5 & 27.0 & 8.4 & 6.5 & 21.0 & 24.5 & 5.5 & 12.0 \\
\quad + Random & \uline{35.2} & 53.3 & \uline{30.0} & 43.3 & 75.0 & \textbf{95.0} & 87.2 & 50.7 & 49.2 & \uline{76.4} & 80.9 & 65.9 & 61.2 \\
\quad + Diversity & 33.8{\textcolor{OrangeRed}{\scriptsize $\downarrow$1.4}} & 53.3 & 28.5{\textcolor{OrangeRed}{\scriptsize $\downarrow$1.5}} & 36.7{\textcolor{OrangeRed}{\scriptsize $\downarrow$6.6}} & 75.0 & 90.0{\textcolor{OrangeRed}{\scriptsize $\downarrow$5.0}} & \uline{87.4}{\textcolor{PineGreen}{\scriptsize $\uparrow$0.2}} & 50.5{\textcolor{OrangeRed}{\scriptsize $\downarrow$0.2}} & 46.7{\textcolor{OrangeRed}{\scriptsize $\downarrow$2.5}} & 74.5{\textcolor{OrangeRed}{\scriptsize $\downarrow$1.9}} & 78.9{\textcolor{OrangeRed}{\scriptsize $\downarrow$2.0}} & \uline{68.1}{\textcolor{PineGreen}{\scriptsize $\uparrow$2.2}} & 60.4{\textcolor{OrangeRed}{\scriptsize $\downarrow$0.8}} \\
\quad + Quality & 30.4{\textcolor{OrangeRed}{\scriptsize $\downarrow$4.8}} & 46.7{\textcolor{OrangeRed}{\scriptsize $\downarrow$6.6}} & 26.3{\textcolor{OrangeRed}{\scriptsize $\downarrow$3.7}} & 30.0{\textcolor{OrangeRed}{\scriptsize $\downarrow$13.3}} & 75.8{\textcolor{PineGreen}{\scriptsize $\uparrow$0.8}} & \textbf{95.0} & 86.4{\textcolor{OrangeRed}{\scriptsize $\downarrow$0.8}} & 48.9{\textcolor{OrangeRed}{\scriptsize $\downarrow$1.8}} & \uline{49.7}{\textcolor{PineGreen}{\scriptsize $\uparrow$0.5}} & 73.0{\textcolor{OrangeRed}{\scriptsize $\downarrow$3.4}} & \uline{83.8}{\textcolor{PineGreen}{\scriptsize $\uparrow$2.9}} & \textbf{72.5}{\textcolor{PineGreen}{\scriptsize $\uparrow$6.6}} & 60.8{\textcolor{OrangeRed}{\scriptsize $\downarrow$0.4}} \\
\quad + Max Loss & 29.6{\textcolor{OrangeRed}{\scriptsize $\downarrow$5.6}} & \uline{56.7}{\textcolor{PineGreen}{\scriptsize $\uparrow$3.4}} & 28.1{\textcolor{OrangeRed}{\scriptsize $\downarrow$1.9}} & \uline{46.7}{\textcolor{PineGreen}{\scriptsize $\uparrow$3.4}} & \uline{77.7}{\textcolor{PineGreen}{\scriptsize $\uparrow$2.7}} & \uline{92.5}{\textcolor{OrangeRed}{\scriptsize $\downarrow$2.5}} & 87.0{\textcolor{OrangeRed}{\scriptsize $\downarrow$0.2}} & \uline{51.6}{\textcolor{PineGreen}{\scriptsize $\uparrow$0.9}} & 44.7{\textcolor{OrangeRed}{\scriptsize $\downarrow$4.5}} & 75.6{\textcolor{OrangeRed}{\scriptsize $\downarrow$0.8}} & 78.6{\textcolor{OrangeRed}{\scriptsize $\downarrow$2.3}} & 67.0{\textcolor{PineGreen}{\scriptsize $\uparrow$1.1}} & 60.0{\textcolor{OrangeRed}{\scriptsize $\downarrow$1.2}} \\
\quad + IFD & 34.6{\textcolor{OrangeRed}{\scriptsize $\downarrow$0.6}} & 54.3{\textcolor{PineGreen}{\scriptsize $\uparrow$1.0}} & 29.4{\textcolor{OrangeRed}{\scriptsize $\downarrow$0.6}} & \uline{46.7}{\textcolor{PineGreen}{\scriptsize $\uparrow$3.4}} & 76.9{\textcolor{PineGreen}{\scriptsize $\uparrow$1.9}} & \textbf{95.0} & 86.4{\textcolor{OrangeRed}{\scriptsize $\downarrow$0.8}} & 50.7 & \uline{49.7}{\textcolor{PineGreen}{\scriptsize $\uparrow$0.5}} & 75.6{\textcolor{OrangeRed}{\scriptsize $\downarrow$0.8}} & 82.9{\textcolor{PineGreen}{\scriptsize $\uparrow$2.0}} & \uline{68.1}{\textcolor{PineGreen}{\scriptsize $\uparrow$2.2}} & \uline{61.6}{\textcolor{PineGreen}{\scriptsize $\uparrow$0.4}} \\
\rowcolor{skyblue}\quad + \textbf{\mymethod{}} (Ours) & \textbf{37.5}{\textcolor{PineGreen}{\scriptsize $\uparrow$2.3}} & \textbf{63.3}{\textcolor{PineGreen}{\scriptsize $\uparrow$10.0}} & \textbf{31.5}{\textcolor{PineGreen}{\scriptsize $\uparrow$1.5}} & \textbf{50.0}{\textcolor{PineGreen}{\scriptsize $\uparrow$6.7}} & \textbf{80.3}{\textcolor{PineGreen}{\scriptsize $\uparrow$5.3}} & \textbf{95.0} & \textbf{88.4}{\textcolor{PineGreen}{\scriptsize $\uparrow$1.2}} & \textbf{53.9}{\textcolor{PineGreen}{\scriptsize $\uparrow$3.2}} & \textbf{55.3}{\textcolor{PineGreen}{\scriptsize $\uparrow$6.1}} & \textbf{78.2}{\textcolor{PineGreen}{\scriptsize $\uparrow$1.8}} & \textbf{84.3}{\textcolor{PineGreen}{\scriptsize $\uparrow$3.4}} & 71.4{\textcolor{PineGreen}{\scriptsize $\uparrow$6.6}} & \textbf{64.5}{\textcolor{PineGreen}{\scriptsize $\uparrow$3.5}} \\

\quad \textit{Full-dataset} & 48.1 & 73.3 & 35.6 & 56.7 & 81.9 & 95.0 & 88.2 & 55.3 & 39.2 & 77.9 & 75.2 & 68.1 & 63.1 \\  \midrule
\multicolumn{14}{l}{\textbf{Qwen2.5-Math-7B-Instruct}} \\
\quad \textit{Zero-shot} & 11.7 & 20.0 & 10.8 & 20.0 & 56.7 & 72.5 & 84.0 & 40.7 & 44.7 & 64.4 & 75.5 & 61.5 & 50.0 \\
\quad + Random & \uline{27.1} & \textbf{50.0} & \uline{22.9} & \textbf{36.7} & \textbf{68.1} & \textbf{92.5} & 84.8 & 47.6 & 48.2 & 70.4 & 82.6 & 68.1 & 57.8 \\
\quad + Diversity & 24.2{\textcolor{OrangeRed}{\scriptsize $\downarrow$2.9}} & 36.7{\textcolor{OrangeRed}{\scriptsize $\downarrow$13.3}} & 21.5{\textcolor{OrangeRed}{\scriptsize $\downarrow$1.4}} & 30.0{\textcolor{OrangeRed}{\scriptsize $\downarrow$6.7}} & 66.9{\textcolor{OrangeRed}{\scriptsize $\downarrow$1.2}} & 85.0{\textcolor{OrangeRed}{\scriptsize $\downarrow$7.5}} & 84.4{\textcolor{OrangeRed}{\scriptsize $\downarrow$0.4}} & 49.2{\textcolor{PineGreen}{\scriptsize $\uparrow$1.6}} & 49.2{\textcolor{PineGreen}{\scriptsize $\uparrow$1.0}} & \uline{71.9}{\textcolor{PineGreen}{\scriptsize $\uparrow$1.5}} & 80.9{\textcolor{OrangeRed}{\scriptsize $\downarrow$1.7}} & 71.9{\textcolor{PineGreen}{\scriptsize $\uparrow$3.8}} & 57.8 \\
\quad + Quality & 23.5{\textcolor{OrangeRed}{\scriptsize $\downarrow$3.6}} & 36.7{\textcolor{OrangeRed}{\scriptsize $\downarrow$13.3}} & 19.2{\textcolor{OrangeRed}{\scriptsize $\downarrow$3.7}} & 20.0{\textcolor{OrangeRed}{\scriptsize $\downarrow$16.7}} & 68.9{\textcolor{PineGreen}{\scriptsize $\uparrow$0.8}} & 87.5{\textcolor{OrangeRed}{\scriptsize $\downarrow$5.0}} & 83.4{\textcolor{OrangeRed}{\scriptsize $\downarrow$1.4}} & \uline{50.7}{\textcolor{PineGreen}{\scriptsize $\uparrow$3.1}} & 48.7{\textcolor{PineGreen}{\scriptsize $\uparrow$0.5}} & 71.7{\textcolor{PineGreen}{\scriptsize $\uparrow$1.3}} & 82.3{\textcolor{OrangeRed}{\scriptsize $\downarrow$0.3}} & 72.5{\textcolor{PineGreen}{\scriptsize $\uparrow$4.4}} & 57.9{\textcolor{PineGreen}{\scriptsize $\uparrow$0.1}} \\
\quad + Max Loss & \textbf{27.3}{\textcolor{PineGreen}{\scriptsize $\uparrow$0.2}} & \uline{46.7}{\textcolor{OrangeRed}{\scriptsize $\downarrow$3.3}} & 21.5{\textcolor{OrangeRed}{\scriptsize $\downarrow$1.4}} & 33.3{\textcolor{OrangeRed}{\scriptsize $\downarrow$3.4}} & 70.0{\textcolor{PineGreen}{\scriptsize $\uparrow$1.9}} & \textbf{92.5} & 85.0{\textcolor{PineGreen}{\scriptsize $\uparrow$0.2}} & 51.0{\textcolor{PineGreen}{\scriptsize $\uparrow$3.4}} & 41.7{\textcolor{OrangeRed}{\scriptsize $\downarrow$6.5}} & 68.6{\textcolor{OrangeRed}{\scriptsize $\downarrow$1.8}} & 79.5{\textcolor{OrangeRed}{\scriptsize $\downarrow$3.1}} & 68.1 & 57.0{\textcolor{OrangeRed}{\scriptsize $\downarrow$0.8}} \\
\quad + IFD & 25.8{\textcolor{OrangeRed}{\scriptsize $\downarrow$1.3}} & 43.3{\textcolor{OrangeRed}{\scriptsize $\downarrow$6.7}} & 22.3{\textcolor{OrangeRed}{\scriptsize $\downarrow$0.6}} & 30.0{\textcolor{OrangeRed}{\scriptsize $\downarrow$6.7}} & 68.3{\textcolor{PineGreen}{\scriptsize $\uparrow$0.2}} & 85.0{\textcolor{OrangeRed}{\scriptsize $\downarrow$7.5}} & \textbf{87.0}{\textcolor{PineGreen}{\scriptsize $\uparrow$2.2}} & 47.6 & 46.7{\textcolor{OrangeRed}{\scriptsize $\downarrow$1.5}} & \uline{71.9}{\textcolor{PineGreen}{\scriptsize $\uparrow$1.5}} & 80.3{\textcolor{OrangeRed}{\scriptsize $\downarrow$2.3}} & 67.0{\textcolor{OrangeRed}{\scriptsize $\downarrow$1.1}} & \uline{59.7}{\textcolor{PineGreen}{\scriptsize $\uparrow$1.9}} \\
\rowcolor{skyblue}\quad + \textbf{\mymethod{}} (Ours) & \textbf{27.3}{\textcolor{PineGreen}{\scriptsize $\uparrow$0.2}} & \uline{46.7}{\textcolor{OrangeRed}{\scriptsize $\downarrow$3.3}} & \textbf{23.1}{\textcolor{PineGreen}{\scriptsize $\uparrow$0.2}} & \textbf{36.7} & \textbf{68.1} & \textbf{92.5} & \textbf{87.0}{\textcolor{PineGreen}{\scriptsize $\uparrow$2.2}} & \textbf{50.1}{\textcolor{PineGreen}{\scriptsize $\uparrow$2.5}} & \textbf{56.8}{\textcolor{PineGreen}{\scriptsize $\uparrow$8.6}} & \textbf{74.0}{\textcolor{PineGreen}{\scriptsize $\uparrow$3.6}} & \textbf{84.3}{\textcolor{PineGreen}{\scriptsize $\uparrow$1.7}} & \textbf{73.6}{\textcolor{PineGreen}{\scriptsize $\uparrow$5.5}} & \textbf{60.5}{\textcolor{PineGreen}{\scriptsize $\uparrow$2.7}} \\
\quad \textit{Full-dataset} & 44.2 & 66.3 & 35.0 & 50.0 & 79.2 & 95.0 & 89.0 & 51.4 & 36.2 & 80.5 & 77.5 & 62.6 & 61.7 \\  \bottomrule 
\multicolumn{14}{l}{\textbf{Llama-3.1-8B-Instruct}} \\
\quad \textit{Zero-shot} & 4.4 & 13.3 & 0.2 & 0.0 & 20.2 & 35.0 & 45.0 & 3.6 & 5.5 & 34.0 & 8.8 & 11.0 & 14.7 \\
\quad + Random & 3.5 & 10.0 & 6.2 & 10.0 & 29.8 & 47.5 & 54.2 & 21.2 & 15.6 & 49.9 & 55.6 & 44.0 & 31.1 \\
\quad + Diversity & 3.7{\textcolor{PineGreen}{\scriptsize $\uparrow$0.2}} & 6.7{\textcolor{OrangeRed}{\scriptsize $\downarrow$3.3}} & 9.4{\textcolor{PineGreen}{\scriptsize $\uparrow$3.2}} & 20.0{\textcolor{PineGreen}{\scriptsize $\uparrow$10.0}} & 29.8 & 40.0{\textcolor{OrangeRed}{\scriptsize $\downarrow$7.5}} & \textbf{58.4}{\textcolor{PineGreen}{\scriptsize $\uparrow$4.2}} & 23.6{\textcolor{PineGreen}{\scriptsize $\uparrow$2.4}} & 16.1{\textcolor{PineGreen}{\scriptsize $\uparrow$0.5}} & 51.9{\textcolor{PineGreen}{\scriptsize $\uparrow$2.0}} & 43.9{\textcolor{OrangeRed}{\scriptsize $\downarrow$11.7}} & 45.1{\textcolor{PineGreen}{\scriptsize $\uparrow$1.1}} & \uline{31.3}{\textcolor{PineGreen}{\scriptsize $\uparrow$0.2}} \\
\quad + Quality & 1.7{\textcolor{OrangeRed}{\scriptsize $\downarrow$1.8}} & 3.3{\textcolor{OrangeRed}{\scriptsize $\downarrow$6.7}} & 4.0{\textcolor{OrangeRed}{\scriptsize $\downarrow$2.2}} & 0.0{\textcolor{OrangeRed}{\scriptsize $\downarrow$10.0}} & 21.1{\textcolor{OrangeRed}{\scriptsize $\downarrow$8.7}} & 35.0{\textcolor{OrangeRed}{\scriptsize $\downarrow$12.5}} & 46.4{\textcolor{OrangeRed}{\scriptsize $\downarrow$7.8}} & 15.9{\textcolor{OrangeRed}{\scriptsize $\downarrow$5.3}} & 3.5{\textcolor{OrangeRed}{\scriptsize $\downarrow$12.1}} & 43.4{\textcolor{OrangeRed}{\scriptsize $\downarrow$6.5}} & 27.1{\textcolor{OrangeRed}{\scriptsize $\downarrow$28.5}} & 19.8{\textcolor{OrangeRed}{\scriptsize $\downarrow$24.2}} & 20.3{\textcolor{OrangeRed}{\scriptsize $\downarrow$10.8}} \\
\quad + Max Loss & 3.1{\textcolor{OrangeRed}{\scriptsize $\downarrow$0.4}} & 6.7{\textcolor{OrangeRed}{\scriptsize $\downarrow$3.3}} & 1.7{\textcolor{OrangeRed}{\scriptsize $\downarrow$4.5}} & 3.3{\textcolor{OrangeRed}{\scriptsize $\downarrow$6.7}} & 23.9{\textcolor{OrangeRed}{\scriptsize $\downarrow$5.9}} & 45.0{\textcolor{OrangeRed}{\scriptsize $\downarrow$2.5}} & 48.8{\textcolor{OrangeRed}{\scriptsize $\downarrow$5.4}} & 18.5{\textcolor{OrangeRed}{\scriptsize $\downarrow$2.7}} & 13.1{\textcolor{OrangeRed}{\scriptsize $\downarrow$2.5}} & 42.3{\textcolor{OrangeRed}{\scriptsize $\downarrow$7.6}} & 45.6{\textcolor{OrangeRed}{\scriptsize $\downarrow$10.0}} & 22.0{\textcolor{OrangeRed}{\scriptsize $\downarrow$22.0}} & 24.3{\textcolor{OrangeRed}{\scriptsize $\downarrow$6.8}} \\
\quad + IFD & 2.5{\textcolor{OrangeRed}{\scriptsize $\downarrow$1.0}} & 13.3{\textcolor{PineGreen}{\scriptsize $\uparrow$3.3}} & 3.5{\textcolor{OrangeRed}{\scriptsize $\downarrow$2.7}} & 13.3{\textcolor{PineGreen}{\scriptsize $\uparrow$3.3}} & 29.4{\textcolor{OrangeRed}{\scriptsize $\downarrow$0.4}} & 47.5 & 50.4{\textcolor{OrangeRed}{\scriptsize $\downarrow$3.8}} & 21.6{\textcolor{PineGreen}{\scriptsize $\uparrow$0.4}} & 17.1{\textcolor{PineGreen}{\scriptsize $\uparrow$1.5}} & 48.8{\textcolor{OrangeRed}{\scriptsize $\downarrow$1.1}} & \uline{58.7}{\textcolor{PineGreen}{\scriptsize $\uparrow$3.1}} & \textbf{50.5}{\textcolor{PineGreen}{\scriptsize $\uparrow$6.5}} & 31.4{\textcolor{PineGreen}{\scriptsize $\uparrow$0.3}} \\
\rowcolor{skyblue}\quad + \textbf{\mymethod{}} (Ours) & \textbf{4.2}{\textcolor{PineGreen}{\scriptsize $\uparrow$0.7}} & \textbf{13.3}{\textcolor{PineGreen}{\scriptsize $\uparrow$3.3}} & \textbf{6.9}{\textcolor{PineGreen}{\scriptsize $\uparrow$0.7}} & \textbf{13.3}{\textcolor{PineGreen}{\scriptsize $\uparrow$3.3}} & \textbf{31.4}{\textcolor{PineGreen}{\scriptsize $\uparrow$1.4}} & \textbf{52.5}{\textcolor{PineGreen}{\scriptsize $\uparrow$5.0}} & \uline{54.6}{\textcolor{PineGreen}{\scriptsize $\uparrow$0.4}} & \textbf{25.3}{\textcolor{PineGreen}{\scriptsize $\uparrow$4.1}} & \textbf{17.1}{\textcolor{PineGreen}{\scriptsize $\uparrow$1.5}} & \textbf{54.5}{\textcolor{PineGreen}{\scriptsize $\uparrow$4.6}} & \textbf{62.4}{\textcolor{PineGreen}{\scriptsize $\uparrow$6.8}} & \uline{44.0} & \textbf{33.4}{\textcolor{PineGreen}{\scriptsize $\uparrow$2.3}} \\
\quad \textit{Full-dataset} & 0.0 & 0.0 & 0.2 & 0.0 & 2.0 & 5.0 & 1.0 & 0.9 & 2.5 & 2.9 & 15.1 & 4.4 & 3.2 \\
\bottomrule 
\multicolumn{14}{l}{\textbf{Llama-3.2-3B}} \\
\quad \textit{Zero-shot} & 0.0 & 0.0 & 0.0 & 0.0 & 0.3& 0.0& 0.2 & 0.0 & 0.0 & 0.8 & 0.0 & 0.0 & 0.2\\
\quad + Random & 0.2 & 0.0 & \textbf{0.4} & 0.0 & 9.4 & \uline{22.5} & 25.6 & 6.5 & 3.0 & \textbf{28.3} & 16.8 & 13.2 & 11.5 \\
\quad + Diversity & 0.4{\textcolor{PineGreen}{\scriptsize $\uparrow$0.2}} & 0.0 & \textbf{0.4} & \textbf{3.3}{\textcolor{PineGreen}{\scriptsize $\uparrow$3.3}} & 10.2{\textcolor{PineGreen}{\scriptsize $\uparrow$0.8}} & 20.0{\textcolor{OrangeRed}{\scriptsize $\downarrow$2.5}} & 24.6{\textcolor{OrangeRed}{\scriptsize $\downarrow$1.0}} & 7.0{\textcolor{PineGreen}{\scriptsize $\uparrow$0.5}} & 2.5{\textcolor{OrangeRed}{\scriptsize $\downarrow$0.5}} & 25.7{\textcolor{OrangeRed}{\scriptsize $\downarrow$2.6}} & 13.4{\textcolor{OrangeRed}{\scriptsize $\downarrow$3.4}} & 4.4{\textcolor{OrangeRed}{\scriptsize $\downarrow$8.8}} & 9.9{\textcolor{OrangeRed}{\scriptsize $\downarrow$1.6}} \\
\quad + Quality & 0.0{\textcolor{OrangeRed}{\scriptsize $\downarrow$0.2}} & 0.0 & 0.0{\textcolor{OrangeRed}{\scriptsize $\downarrow$0.4}} & 0.0 & 6.6{\textcolor{OrangeRed}{\scriptsize $\downarrow$2.8}} & 7.5{\textcolor{OrangeRed}{\scriptsize $\downarrow$15.0}}  & 22.4{\textcolor{OrangeRed}{\scriptsize $\downarrow$3.2}} & 4.6{\textcolor{OrangeRed}{\scriptsize $\downarrow$1.9}} & \textbf{4.5}{\textcolor{PineGreen}{\scriptsize $\uparrow$1.5}} & 24.4{\textcolor{OrangeRed}{\scriptsize $\downarrow$3.9}} & \textbf{21.9}{\textcolor{PineGreen}{\scriptsize $\uparrow$5.1}} & 13.2 & 10.8{\textcolor{OrangeRed}{\scriptsize $\downarrow$0.7}} \\
\quad + Max Loss & 0.2 & \textbf{3.3}{\textcolor{PineGreen}{\scriptsize $\uparrow$3.3}} & 0.0{\textcolor{OrangeRed}{\scriptsize $\downarrow$0.4}} & 0.0 & 6.1{\textcolor{OrangeRed}{\scriptsize $\downarrow$3.3}} & 10.0{\textcolor{OrangeRed}{\scriptsize $\downarrow$12.5}} & 23.4{\textcolor{OrangeRed}{\scriptsize $\downarrow$2.2}} & 5.5{\textcolor{OrangeRed}{\scriptsize $\downarrow$1.0}} & \uline{4.0}{\textcolor{PineGreen}{\scriptsize $\uparrow$1.0}} & 25.7{\textcolor{OrangeRed}{\scriptsize $\downarrow$2.6}} & \uline{18.5}{\textcolor{PineGreen}{\scriptsize $\uparrow$1.7}} & 11.0{\textcolor{OrangeRed}{\scriptsize $\downarrow$2.2}} & 10.5{\textcolor{OrangeRed}{\scriptsize $\downarrow$1.0}} \\
\quad + IFD & \textbf{1.0}{\textcolor{PineGreen}{\scriptsize $\uparrow$0.8}} & 0.0 & 0.2{\textcolor{OrangeRed}{\scriptsize $\downarrow$0.2}} & 0.0 & 9.4 & \uline{22.5} & 25.6 & 6.8{\textcolor{PineGreen}{\scriptsize $\uparrow$0.3}} & 0.5{\textcolor{OrangeRed}{\scriptsize $\downarrow$2.5}} & 24.4{\textcolor{OrangeRed}{\scriptsize $\downarrow$3.9}} & 13.1{\textcolor{OrangeRed}{\scriptsize $\downarrow$3.7}} & 6.6{\textcolor{OrangeRed}{\scriptsize $\downarrow$6.6}} & \uline{9.7}{\textcolor{OrangeRed}{\scriptsize $\downarrow$1.8}} \\
\rowcolor{skyblue}\quad + \textbf{\mymethod{}} (Ours) & \textbf{1.0}{\textcolor{PineGreen}{\scriptsize $\uparrow$0.8}} & 0.0 & \textbf{0.4} & \textbf{3.3}{\textcolor{PineGreen}{\scriptsize $\uparrow$3.3}} & \textbf{12.3}{\textcolor{PineGreen}{\scriptsize $\uparrow$2.9}} & \textbf{27.5}{\textcolor{PineGreen}{\scriptsize $\uparrow$5.0}} & \textbf{27.8}{\textcolor{PineGreen}{\scriptsize $\uparrow$2.2}} & \textbf{8.6}{\textcolor{PineGreen}{\scriptsize $\uparrow$2.1}} & \textbf{4.5}{\textcolor{PineGreen}{\scriptsize $\uparrow$1.5}} & \uline{27.3}{\textcolor{OrangeRed}{\scriptsize $\downarrow$1.0}} & 17.1{\textcolor{PineGreen}{\scriptsize $\uparrow$0.3}} & \textbf{15.4}{\textcolor{PineGreen}{\scriptsize $\uparrow$2.2}} & \textbf{12.7}{\textcolor{PineGreen}{\scriptsize $\uparrow$1.2}} \\
\quad \textit{Full-dataset} & 5.0 & 13.3 & 9.0 & 30.0 & 36.4 & 52.5 & 54.2 & 24.3 & 2.5 & 36.4 & 38.7 & 24.2 & 25.6 \\  \midrule
\end{tabular}
}
\end{table*}

%% file: chapters/4_exp.tex
\section{Experiments}
\label{sec:exp}

\subsection{Experimental Setup}
\noindent \textbf{Training Details.} 
We conducted experiments on the OpenR1-Math-220K dataset, synthesized by DeepSeek-R1 for long CoT training~\citep{deepseekai2025deepseekr1incentivizingreasoningcapability}. 
After filtering for samples with correct reasoning traces, we obtained about 196k high-quality examples. 
As the reference model for our data scoring procedure (detailed in Sections~\ref{sec:heads} and~\ref{sec:scores}), we employed {Qwen2.5-Math-1.5B-Instruct}~\citep{yang2024qwen25math}, which complemented a weak-to-strong function. 
We validated our method's effectiveness and generalizability by fine-tuning four diverse large language models: Qwen2.5-Math-7B-Base~\citep{yang2024qwen25math}, Qwen2.5-Math-7B-Instruct~\citep{yang2024qwen25math}, Llama-3.2-3B~\citep{llama3_2} and Llama-3.1-8B-Instruct~\citep{llama3_1}. 
For detailed training parameters, please refer to Appendix~\ref{app:detailed_setting}.

\noindent \textbf{Baselines.} We compared our method with several representative data selection strategies: 
(i) Random: Randomly selected samples from the training data pool; 
(ii) Max Loss: Computed the training loss of each sample using Qwen2.5-Math-1.5B-Instruct and selected the samples with the highest loss; 
(iii) Quality: This baseline uses Qwen2.5-Math-1.5B-Instruct to evaluate data quality via prompt-based scoring and selects the top-rated samples.
(iv) Diversity~\citep{muennighoff2025s1simpletesttimescaling}: Data are uniformly sampled from different categories of mathematical problems to ensure diversity;
(v) IFD~\citep{li-etal-2024-quantity}: Instruction-Following Difficulty is defined as the ratio between the conditioned answer loss (with instruction) and the answer loss (without instruction), and we selected the samples with the largest instruction-following difficulty.

\noindent \textbf{Evaluation Settings.} 
We used a rule-based evaluation framework~\citep{ye2025limoreasoning}, including metrics such as Pass@1 (P@1)~\citep{chen2021codex} and Majority@16 (M@16). 
We evaluated on nine benchmarks, including three competition-level benchmarks: AIME 2024, AIME 2025, and AMC 2023, and six datasets such as MATH500~\citep{hendrycks2021math}, OlympiadBench~\citep{he2024olympiadbench}, GAOKAO 2023 \& 2024~\citep{yang2024qwen25math}, GAOKAO MATH~\citep{yang2024qwen25math}, and KAOYAN~\citep{ye2025limoreasoning}. 
The system prompt for evaluation was: \textit{Please reason step by step, and put your final answer within \textbackslash boxed\{\}}. 
For competition benchmarks, 16 solutions per problem were sampled. 
For other benchmarks, we used greedy decoding and sampled one solution per problem.

\begin{figure*}[tb!]
    \centering
    \vspace{-5pt}
    \includegraphics[width=\textwidth]{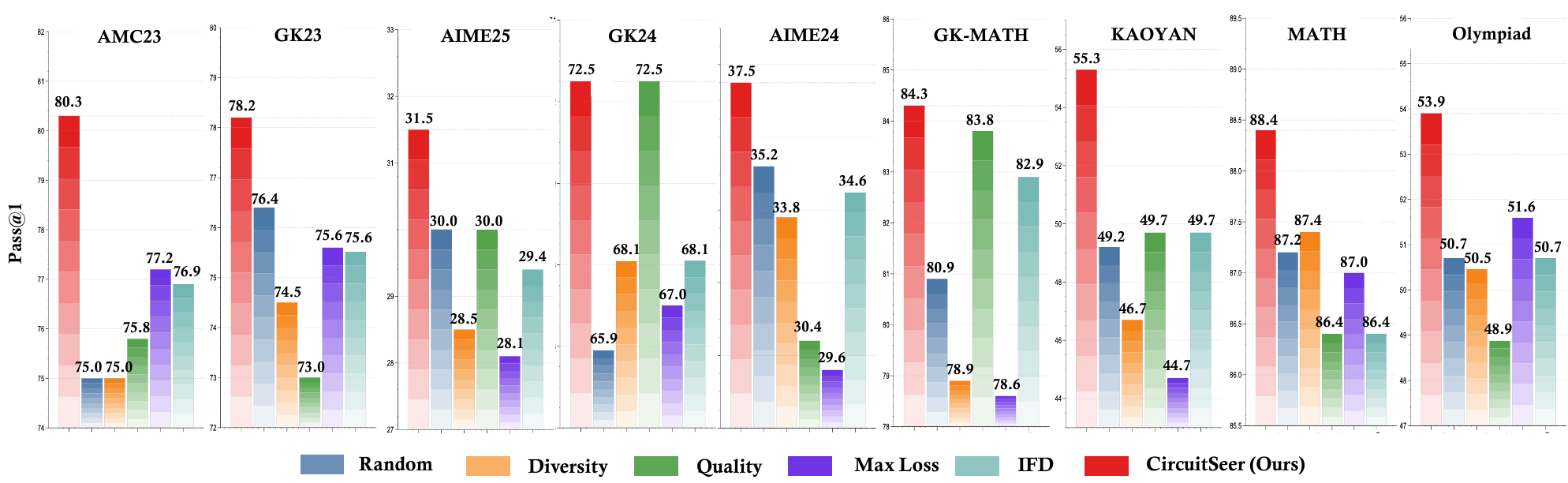}
    \caption{Pass@1 Performance comparison of CircuitSeer against baseline methods across 9 benchmark datasets on Qwen2.5-Math-7B-Base. Our method consistently outperforms all baselines, demonstrating the effectiveness of leveraging internal reasoning circuits for data selection.}
    \vspace{-5pt}
    \label{fig:performance_comparison}
\end{figure*}

\subsection{Main Results}
\label{sec:Main Results}

\paragraph{CircuitSeer Outperformed All Baseline Methods.}
As illustrated in Table~\ref{tab:main_results_detailed}, CircuitSeer consistently surpassed all baseline data selection techniques across a suite of nine benchmarks, showcasing reliable and consistent enhancements. On the Qwen2.5-Math-7B-Base model, CircuitSeer achieved Pass@1 (P@1) scores of 37.5 on AIME24 and 31.5 on AIME25, notably outperforming Random Sampling and yielding substantial improvements of 10.0 and 6.7 points on the Majority@16 (M@16) metric. On AMC23, CircuitSeer excelled with a P@1 of 80.3, exceeding the best alternative by 2.6 points, while on Kaoyan, it reached 55.3, marking a 6.1-point advancement over Random. In the Qwen2.5-Math-7B-Instruct model, CircuitSeer achieved notable progress, with an impressive gain of 8.6 points on Kaoyan and a 5.5-point increase on GK24, underscoring its adaptability across model variations. Even when deployed on the smaller Llama-3.2-3B, CircuitSeer still delivered remarkable improvements, raising AMC23's M@16 metric to 27.5, a 5.0-point enhancement, and boosting MATH's P@1 to 27.8, reflecting a 2.2-point increase. These results collectively demonstrated CircuitSeer's improvements in mathematical reasoning across benchmarks.
\paragraph{Generalization Across Diverse Models.}
CircuitSeer consistently demonstrated generalization across models of varying capacities. On the Qwen2.5-Math-7B-Instruct model, CircuitSeer excelled by achieving a P@1 of 27.3 on AIME24, surpassing other methods like Diversity and Quality, which achieved 24.2 and 23.5, respectively. On KAOYAN, CircuitSeer reached 56.8, outperforming the next best method, Max Loss, which attained 41.7. Similarly, on GK24, CircuitSeer achieved 73.6, marking a significant improvement over the best baseline method, Quality, which scored 72.5. For the Qwen2.5-Math-7B-Base model, CircuitSeer yielded remarkable improvements, with P@1 on AIME25 rising to 31.5, compared to the next best method, Random, which scored 30.0. On MATH, CircuitSeer achieved 88.4, surpassing Diversity's 87.4, and on Kaoyan, it attained 55.3, which was significantly higher than Random's 49.2. Even with the smaller Llama-3.2-3B model, CircuitSeer delivered substantial improvements, such as AMC23 M@16 increasing to 27.5, outperforming Diversity, which achieved 20.0, and MATH P@1 reaching 27.8, compared to Random's 25.6.

\subsection{Ablation Studies}
\label{sec:ablation}

To validate the efficacy of our proposed components and understand their individual contributions, we conducted a series of ablation studies on the Qwen2.5-Math-7B. Our analysis was designed to answer four critical questions: (1) Was the identification of specialized reasoning heads essential? (2) Did our proposed score reflect the value of the reasoning data? (3) What was the impact of our soft sampling strategy compared to Top-K selection? (4) Were reasoning signals better captured from the input or the output of the sample?

\paragraph{Impact of Head Selection}
A fundamental aspect of our approach was the identification of a specialized subset of attention heads crucial for mathematical reasoning, which offered a clearer and more pertinent signal for data selection compared to using all heads or a random subset. To validate this, we assessed three strategies for score computation:
\begin{enumerate}[leftmargin=10pt, topsep=0pt, itemsep=1pt, partopsep=1pt, parsep=1pt]
\vspace{0.25em}
    \item \textbf{Reasoning Heads}: Utilized the reasoning heads identified through our intervention-based method detailed in Section~\ref{sec:heads}.
    \item \textbf{All Heads}: Calculated scores based on the attention weights from all heads in the model.
    \item \textbf{Random Heads}: Selected a random subset of $k$ heads (matching the number used in our method) as the basis for scoring.
\end{enumerate}

\input{tables/ablation_heads}

As presented in Table \ref{tab:ablation_heads_comprehensive}, the data subset selected with our identified reasoning heads yielded substantially superior performance. This result confirmed that these heads were indeed specialized for processing structural and logical information within reasoning traces. In contrast, using all heads introduced significant noise from heads dedicated to other, irrelevant functions (e.g., positional or syntactic processing), which diluted the quality signal and degraded selection performance. Unsurprisingly, random heads lacked the necessary signal altogether, performing only marginally better than random data selection, which underscored the necessity of our systematic head identification.

\input{tables/ablation_scores}

\input{tables/ablation_sample}

\paragraph{Effectiveness of CircuitSeer Score.}
As shown in Table~\ref{tab:main_results_detailed}, soft sampling guided by high CircuitSeer scores consistently achieves state-of-the-art performance across multiple models.
To validate these scores, we conducted a contrastive experiment comparing models trained on high- versus low-score subsets (Table~\ref{tab:ablation_score_strategy}).
Results reveal a clear trend: high-score samples involve problems requiring deeper multi-step reasoning, while low-score ones mostly capture simpler questions (see Appendix~\ref{app:case_study}). 
This performance gap confirms that the CircuitSeer score effectively captures the structural and instructional quality of mathematical reasoning data, serving as a reliable signal for data selection.

\paragraph{Soft Sampling vs. Top-k Selection.}
We examined our sampling strategy. Although Top-k selection is straightforward, it tends to over-concentrate on redundant high-score samples, reducing diversity. In contrast, our score-guided Soft Sampling maintains quality while introducing stochasticity for broader coverage. We compared it against a deterministic baseline that directly selected the highest-scoring samples.

As shown in Table \ref{tab:ablation_sampling_strategy}, soft sampling achieved superior performance. We attributed this to its ability to foster a more diverse training subset by occasionally including lower-scoring-but-still-valuable samples. This stochasticity prevented the selection process from collapsing into a narrow region of the data distribution, which enhanced the model's generalization capabilities on unseen problems.

\input{tables/ablation_output}
\paragraph{Input vs. Output Scoring.} 
We examined whether reasoning signals were better captured from the problem statement or from the generated reasoning trajectory by comparing two variants: (1) output-based scoring, which assigned scores based on the solutions of the training samples, and (2) input-based scoring, which focused solely on the problem inputs of the training samples.
As shown in Table~\ref{tab:input_output_ablation}, the input-based variant outperformed the output-based one across 7 out of 9 benchmarks, achieving an average Pass@1 (P@1) of 64.5\% versus 62.8\%. 
Input-based scoring captured reasoning complexity more effectively because attention patterns during the input phase reflected how the model parsed the problem, identified constraints, and constructed internal logical structures—revealing \textit{what reasoning is required}. 
In contrast, activations during the output phase primarily reflected how reasoning is expressed, being shaped by stylistic factors such as solution format, generation length, and linguistic style rather than true reasoning depth.

\subsection{Analysis}

\begin{figure}[tb!]
    \centering
    \includegraphics[width=\columnwidth]{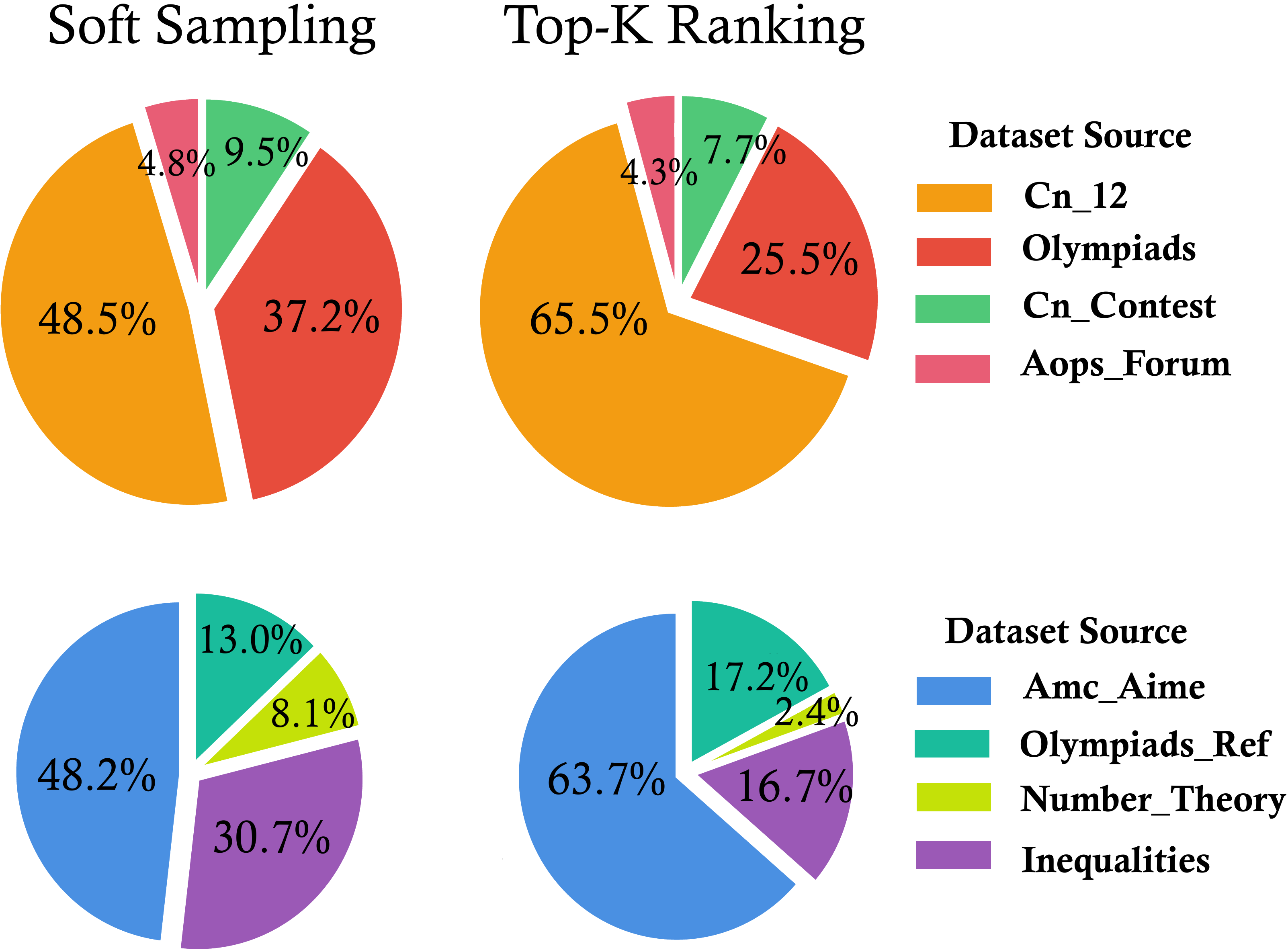}
    \vspace{-5pt} 
    \caption{
    Comparison of the source data distribution for subsets selected by Soft Sampling versus the Top-k Selection baseline. To enhance visual clarity, less frequent categories (\texttt{Amc\_Aime}, \texttt{Inequalities}, \texttt{Olympiads\_Ref}, and \texttt{Number\_Theory}) are aggregated into a single group, accounting for 2.51\% and 2.92\% of the Top-k and Soft Sampling subsets, respectively.
}
    \vspace{-8pt}
    \label{fig:category_distribution}
\end{figure}

\paragraph{Soft Sampling Fosters Superior Data Diversity.}
To understand why our Soft Sampling approach consistently outperforms the Top‑k baseline (Table~\ref{tab:ablation_sampling_strategy}), we first examined the categorical distribution of the selected subsets. As shown in Figure~\ref{fig:category_distribution}, the Top‑K method exhibits a pronounced selection bias, heavily oversampling from the \texttt{Cn\_k12} category. This bias yields a modest advantage on benchmarks of a similar style, such as GAOKAO‑Math and GAOKAO24, but comes at the cost of reduced coverage of other problem types and weaker overall generalization. This trade-off suggests that even a moderate emphasis on diversity can be essential for better generalization.

\begin{figure}[tb!]
    \centering
    \vspace{-5pt}
    \includegraphics[width=\columnwidth]{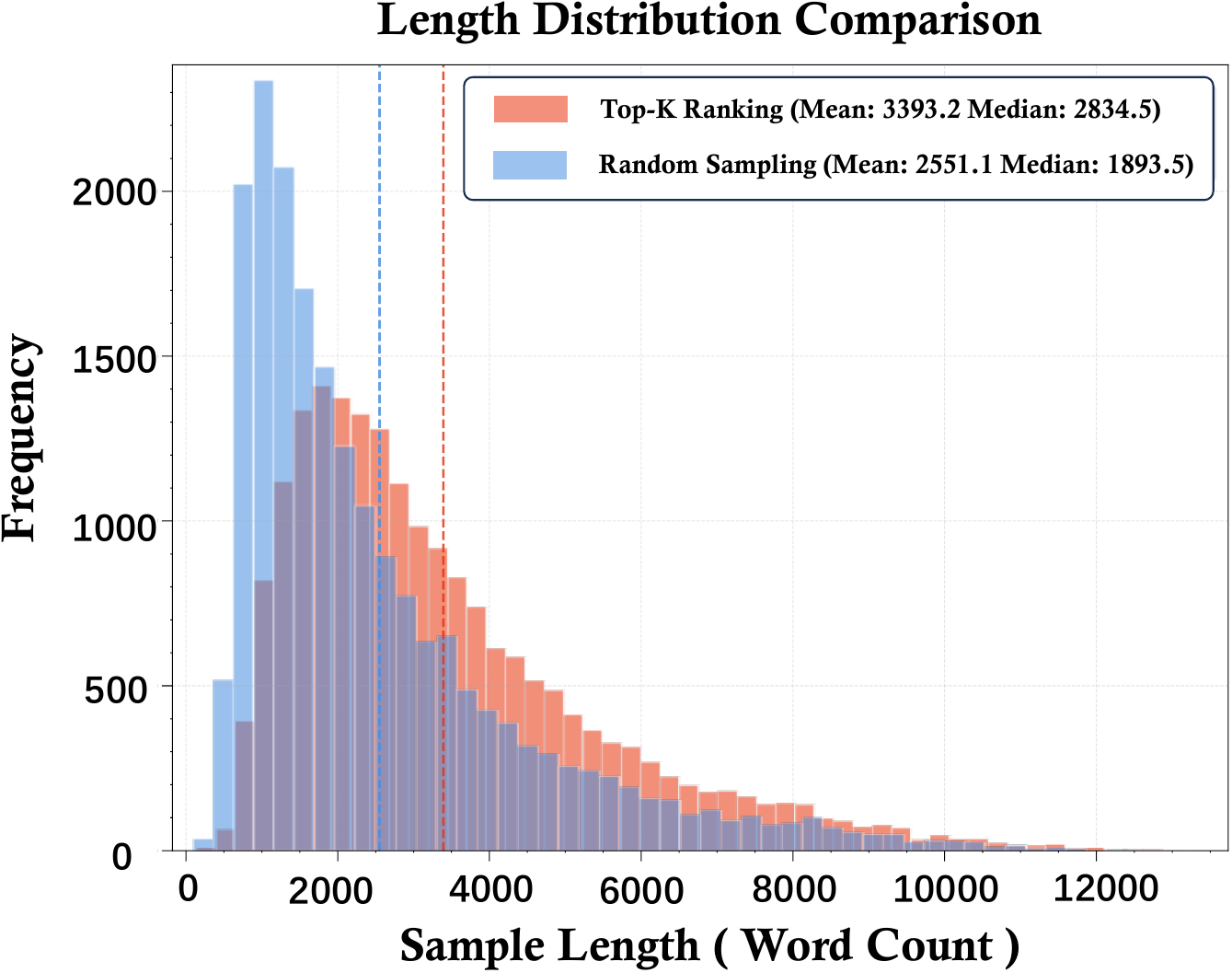} 
    \vspace{-10pt}
    \caption{
        Comparison of the example length distribution between subsets selected by Top-K ranking and Random sampling. Top-K ranking demonstrates a clear preference for longer examples, resulting in significantly higher mean and median lengths, with the respective means marked by dashed lines.
    }

    \label{fig:length_distribution}
    \vspace{-8pt}
\end{figure}

\paragraph{Correlation Between Sample Length and Performance.}
Beyond categorical diversity, we analyze the impact of example length on model performance with respect to the CircuitSeer score. Figure~\ref{fig:length_distribution} shows that Top‑k ranking, by selecting the highest‑scoring samples, strongly favors longer examples compared to a random baseline, with mean and median word counts of 3393.2 and 2834.5 versus 2551.1 and 1893.5, respectively. This aligns with the intuition that longer reasoning traces often encode more complex, multi‑step logic that can be valuable for training. However, a simple preference for length is insufficient: while Top‑k improves average performance (Table~\ref{tab:ablation_sampling_strategy}), it underperforms the random baseline on three of nine benchmarks, likely due to reduced diversity or over‑emphasis on a single reasoning style. In contrast, Soft Sampling leverages the same scores but maintains diversity, achieving superior results across all benchmarks. This reinforces the core insight that effective reasoning data selection must balance quality and diversity to foster robust, broadly applicable models.
\vspace{-5pt}

%% file: tables/ablation_heads.tex
\begin{table}[tb!]
\centering
\vspace{2pt}
\caption{
    Comprehensive ablation study on head selection strategies.
    The efficacy of our proposed Math Heads was validated against two baselines: using All Heads and a Random selection of heads.
    This table presents the change in Pass@1 (P@1) performance for each method relative to the Random Heads baseline.
    All experiments were performed on the Qwen2.5-Math-7B.
}
\label{tab:ablation_heads_comprehensive}
\small
\vspace{-8pt}

\setlength{\tabcolsep}{10pt}
\resizebox{.49\textwidth}{!}{
\begin{tabular}{@{}l|ccc@{}}
\toprule
\textbf{Dataset} & \textbf{Random} & \textbf{All} & \textbf{Reasoning} \\
\midrule
AIME24      & 35.2 & 33.8{\textcolor{OrangeRed}{\scriptsize $\downarrow$1.4}} & 37.5{\textcolor{PineGreen}{\scriptsize $\uparrow$2.3}} \\
AIME25      & 30.0 & 26.0{\textcolor{OrangeRed}{\scriptsize $\downarrow$4.0}} & 31.5{\textcolor{PineGreen}{\scriptsize $\uparrow$1.5}} \\
AMC23       & 75.0 & 75.6{\textcolor{PineGreen}{\scriptsize $\uparrow$0.6}} & 80.3{\textcolor{PineGreen}{\scriptsize $\uparrow$5.3}} \\
MATH        & 87.2 & 87.6{\textcolor{PineGreen}{\scriptsize $\uparrow$0.4}} & 88.4{\textcolor{PineGreen}{\scriptsize $\uparrow$1.2}} \\
Olympiad    & 50.7 & 51.1{\textcolor{PineGreen}{\scriptsize $\uparrow$0.4}} & 53.9{\textcolor{PineGreen}{\scriptsize $\uparrow$3.2}} \\
Kaoyan      & 49.2 & 42.2{\textcolor{OrangeRed}{\scriptsize $\downarrow$7.0}} & 55.3{\textcolor{PineGreen}{\scriptsize $\uparrow$6.1}} \\
GK23        & 76.4 & 78.2{\textcolor{PineGreen}{\scriptsize $\uparrow$1.8}} & 78.2{\textcolor{PineGreen}{\scriptsize $\uparrow$1.8}} \\
GK-Math     & 80.9 & 80.9{\textcolor{OrangeRed}{\scriptsize $\downarrow$0.0}} & 84.3{\textcolor{PineGreen}{\scriptsize $\uparrow$3.4}} \\
GK24        & 65.9 & 64.9{\textcolor{OrangeRed}{\scriptsize $\downarrow$1.0}} & 71.4{\textcolor{PineGreen}{\scriptsize $\uparrow$5.5}} \\
\midrule
Average     & 61.2 & 60.0{\textcolor{OrangeRed}{\scriptsize $\downarrow$1.2}} & 64.5{\textcolor{PineGreen}{\scriptsize $\uparrow$3.3}} \\
\bottomrule
\end{tabular}%
}
\vspace{-5pt}
\end{table}

%% file: tables/ablation_scores.tex
\begin{table}[tb!]
\centering
\vspace{-5pt}
\caption{
   Ablation study on input score-based data selection.
    We compared Input High-Score Sampling with Input Low-Score Sampling and Random Sampling.
    All scores are Pass@1 (P@1) accuracy.
    Selecting high-score samples outperformed other strategies.
}
\label{tab:ablation_score_strategy}
\small
\vspace{-8pt}

\setlength{\tabcolsep}{10pt}
\resizebox{.49\textwidth}{!}{
\begin{tabular}{@{}l|ccc@{}}
\toprule
\textbf{Dataset} & \textbf{Random} & \textbf{Low} & \textbf{High} \\
\midrule
AIME24   & 35.2 & 29.8{\textcolor{OrangeRed}{\scriptsize $\downarrow$5.4}} & 37.5{\textcolor{PineGreen}{\scriptsize $\uparrow$2.3}} \\
AIME25   & 30.0 & 25.9{\textcolor{OrangeRed}{\scriptsize $\downarrow$4.1}} & 31.5{\textcolor{PineGreen}{\scriptsize $\uparrow$1.5}} \\
AMC23    & 75.0 & 70.2{\textcolor{OrangeRed}{\scriptsize $\downarrow$4.8}} & 80.3{\textcolor{PineGreen}{\scriptsize $\uparrow$5.3}} \\
MATH     & 87.2 & 80.1{\textcolor{OrangeRed}{\scriptsize $\downarrow$7.1}} & 88.4{\textcolor{PineGreen}{\scriptsize $\uparrow$1.2}} \\
Olympiad & 50.7 & 45.2{\textcolor{OrangeRed}{\scriptsize $\downarrow$5.5}} & 53.9{\textcolor{PineGreen}{\scriptsize $\uparrow$3.2}} \\
Kaoyan   & 49.2 & 46.2{\textcolor{OrangeRed}{\scriptsize $\downarrow$3.0}} & 55.3{\textcolor{PineGreen}{\scriptsize $\uparrow$6.1}} \\
GK23     & 76.4 & 73.4{\textcolor{OrangeRed}{\scriptsize $\downarrow$3.0}} & 78.2{\textcolor{PineGreen}{\scriptsize $\uparrow$1.8}} \\
GK-Math  & 80.9 & 76.8{\textcolor{OrangeRed}{\scriptsize $\downarrow$4.1}} & 84.3{\textcolor{PineGreen}{\scriptsize $\uparrow$3.4}} \\
GK24     & 65.9 & 65.1{\textcolor{OrangeRed}{\scriptsize $\downarrow$0.8}} & 71.4{\textcolor{PineGreen}{\scriptsize $\uparrow$5.5}} \\
\midrule
Average  & 61.2 & 57.2{\textcolor{OrangeRed}{\scriptsize $\downarrow$4.0}} & 64.5{\textcolor{PineGreen}{\scriptsize $\uparrow$3.3}} \\
\bottomrule
\end{tabular}%
}
\vspace{-8pt}
\end{table}

%% file: tables/ablation_sample.tex
\begin{table}[tb!]
\centering
\vspace{5pt}
\caption{
    Ablation study comparing Soft Sampling and Top-k selection against a Random baseline. Performance changes for both methods are shown relative to Random. All scores are Pass@1 (P@1) accuracy. Soft Sampling demonstrates the most significant average improvement.
}
\label{tab:ablation_sampling_strategy}
\small
\vspace{-8pt}

\setlength{\tabcolsep}{10pt}
\resizebox{.49\textwidth}{!}{%
\begin{tabular}{@{}l|ccc@{}}
\toprule
\textbf{Dataset} & \textbf{Random} & \textbf{Hard (Top-k)} & \textbf{Soft} \\
\midrule
AIME24   & 35.2 & 33.1{\textcolor{OrangeRed}{\scriptsize $\downarrow$2.1}} & 37.5{\textcolor{PineGreen}{\scriptsize $\uparrow$2.3}} \\
AIME25   & 30.0 & 30.8{\textcolor{PineGreen}{\scriptsize $\uparrow$0.8}} & 31.5{\textcolor{PineGreen}{\scriptsize $\uparrow$1.5}} \\
AMC23    & 75.0 & 77.5{\textcolor{PineGreen}{\scriptsize $\uparrow$2.5}} & 80.3{\textcolor{PineGreen}{\scriptsize $\uparrow$5.3}} \\
MATH     & 87.2 & 87.6{\textcolor{PineGreen}{\scriptsize $\uparrow$0.4}} & 88.4{\textcolor{PineGreen}{\scriptsize $\uparrow$1.2}} \\
Olympiad & 50.7 & 49.5{\textcolor{OrangeRed}{\scriptsize $\downarrow$1.2}} & 53.9{\textcolor{PineGreen}{\scriptsize $\uparrow$3.2}} \\
Kaoyan   & 49.2 & 50.8{\textcolor{PineGreen}{\scriptsize $\uparrow$1.6}} & 55.3{\textcolor{PineGreen}{\scriptsize $\uparrow$6.1}} \\
GK23     & 76.4 & 75.3{\textcolor{OrangeRed}{\scriptsize $\downarrow$1.1}} & 78.2{\textcolor{PineGreen}{\scriptsize $\uparrow$1.8}} \\
GK-Math  & 80.9 & 86.9{\textcolor{PineGreen}{\scriptsize $\uparrow$6.0}} & 84.3{\textcolor{PineGreen}{\scriptsize $\uparrow$3.4}} \\
GK24     & 65.9 & 73.6{\textcolor{PineGreen}{\scriptsize $\uparrow$7.7}} & 71.4{\textcolor{PineGreen}{\scriptsize $\uparrow$5.5}} \\
\midrule
Average  & 61.2 & 62.8{\textcolor{PineGreen}{\scriptsize $\uparrow$1.6}} & 64.5{\textcolor{PineGreen}{\scriptsize $\uparrow$3.3}} \\
\bottomrule
\end{tabular}%
}
\vspace{-10pt}
\end{table}

%% file: tables/ablation_output.tex
\begin{table}[tb!]
\centering
\vspace{-4pt}
\caption{
    Ablation study comparing outputs and inputs against a Random baseline. Performance changes for both are shown relative to Random. All scores are Pass@1 (P@1) accuracy.
}
\label{tab:input_output_ablation}
\small
\vspace{-8pt}

\setlength{\tabcolsep}{10pt}
\resizebox{.49\textwidth}{!}{%
\begin{tabular}{@{}l|ccc@{}}
\toprule
\textbf{Dataset} & \textbf{Random} & \textbf{Output} & \textbf{Input} \\
\midrule
AIME24   & 35.2 & 35.6{\textcolor{PineGreen}{\scriptsize $\uparrow$0.4}} & 37.5{\textcolor{PineGreen}{\scriptsize $\uparrow$2.3}} \\
AIME25   & 30.0 & 29.8{\textcolor{OrangeRed}{\scriptsize $\downarrow$0.2}} & 31.5{\textcolor{PineGreen}{\scriptsize $\uparrow$1.5}} \\
AMC23    & 75.0 & 73.9{\textcolor{OrangeRed}{\scriptsize $\downarrow$1.1}} & 80.3{\textcolor{PineGreen}{\scriptsize $\uparrow$5.3}} \\
MATH     & 87.2 & 89.0{\textcolor{PineGreen}{\scriptsize $\uparrow$1.8}} & 88.4{\textcolor{PineGreen}{\scriptsize $\uparrow$1.2}} \\
Olympiad & 50.7 & 54.7{\textcolor{PineGreen}{\scriptsize $\uparrow$4.0}} & 53.9{\textcolor{PineGreen}{\scriptsize $\uparrow$3.2}} \\
Kaoyan   & 49.2 & 48.2{\textcolor{OrangeRed}{\scriptsize $\downarrow$1.0}} & 55.3{\textcolor{PineGreen}{\scriptsize $\uparrow$6.1}} \\
GK23     & 76.4 & 78.2{\textcolor{PineGreen}{\scriptsize $\uparrow$1.8}} & 78.2{\textcolor{PineGreen}{\scriptsize $\uparrow$1.8}} \\
GK-Math  & 80.9 & 85.8{\textcolor{PineGreen}{\scriptsize $\uparrow$4.9}} & 84.3{\textcolor{PineGreen}{\scriptsize $\uparrow$3.4}} \\
GK24     & 65.9 & 70.3{\textcolor{PineGreen}{\scriptsize $\uparrow$4.4}} & 71.4{\textcolor{PineGreen}{\scriptsize $\uparrow$5.5}} \\
\midrule
Average  & 61.2 & 62.8{\textcolor{PineGreen}{\scriptsize $\uparrow$1.6}} & 64.5{\textcolor{PineGreen}{\scriptsize $\uparrow$3.3}} \\
\bottomrule
\end{tabular}%
}
\vspace{-10pt}
\end{table}

%% file: chapters/5_conclusion.tex
\section{Conclusion}
\vspace{-5pt}
In this paper, we present CircuitSeer, a mechanism-driven framework that identifies high-quality reasoning data by probing the internal reasoning circuits of large language models (LLMs). By quantifying activations of reasoning attention heads, CircuitSeer provides an interpretable, model-internal signal of reasoning complexity, effectively eliminating the need for external heuristics. Experiments across diverse models and benchmarks consistently show that fine-tuning on only 10\% of CircuitSeer-selected data outperforms full-data training, demonstrating both efficiency and robustness. Beyond empirical gains, this work highlights that reasoning circuits encode measurable structures of problem difficulty, suggesting that the internal dynamics of LLMs can serve as principled indicators for data quality. This mechanism-aware perspective enables more transparent, scalable, and domain-agnostic data curation, thereby bridging mechanistic interpretability with practical LLMs training.

%% file: chapters/X_appendix.tex
\appendix

\section{Detailed Experimental Settings}
\label{app:detailed_setting}
\noindent \textbf{Training Details.} All models are trained using the llamafactory framework with a learning rate of $5 \times 10^{-5}$ for 3 epochs, utilizing fused AdamW as the optimizer. The total batch size is set to 32, with a warmup ratio of 0.1, and a cosine decay schedule towards zero. Both Flash Attention and DeepSpeed ZeRO-3 are adopted for training optimization.

For the Qwen-2.5-Math-7B model, we extend the model’s context length from 4,096 to 16,384 via RoPE scaling, increasing the RoPE frequency from 10k to 300k. The max training sequence length is set to 16k. For Llama-3.2-3B, the max training sequence length is set to 8k. All experiments are conducted on 8 NVIDIA A100 GPUs. For Qwen2.5-Math-7B-Base, Qwen2.5-Math-7B-Instruct, and Llama-3.1-8B-Instruct, training on the full dataset takes approximately 54 hours, while training on the selected 10\% of the data takes about 6 hours. For Llama-3.2-3B, full-dataset training requires roughly 27 hours, compared to approximately 3 hours for the selected 10\% subset.

For the selection of reasoning heads, the probe dataset for the Detecting Heads stage is constructed using the top 300 samples from the OpenR1-Math-196k (After Filtered) dataset, sorted in descending order by their loss on the reference model. The number of selected top-k heads, k is set to 5\% of the total number of attention heads for each model.

\section{Case Study}
\label{app:case_study}
\input{chapters/case_study}

\subsection{Pseudo Code of CircuitSeer}\label{sec:pseudo_code}

The detailed pseudo code of CircuitSeer is demonstrated in Algorithm~\ref{alg:circuitseer}.

\begin{algorithm*}[tb!]
\caption{CircuitSeer for Reasoning Circuits-Based Data Selection}
\textbf{Input:} \\
$\mathcal{M}_{\mathrm{ref}}$ -- Reference pretrained language model \\
$\mathcal{D}$ -- Full reasoning dataset \\
$\mathcal{D}_{\mathrm{probe}}$ -- Probe dataset for circuit importance evaluation \\
$k$ -- Number of top reasoning heads to select \\
$\rho$ -- Budget ratio for subset size \\
\textbf{Output:} $\mathcal{D'}$ -- Selected subset ($\mathcal{D'} \subseteq \mathcal{D}$, $|\mathcal{D'}| < |\mathcal{D}|$)
\begin{algorithmic}[1]
\Procedure{CircuitSeer}{$\mathcal{M}_{\mathrm{ref}}, \mathcal{D}, \mathcal{D}_{\mathrm{probe}}, k, \rho$}
    \State $\mathcal{H} \gets$ all attention heads in $\mathcal{M}_{\mathrm{ref}}$ \Comment{Initialize head set}
    \State $\mathrm{Imp} \gets \mathbf{0}$ \Comment{Initialize importance scores}
    \For{$h \in \mathcal{H}$}
        \State Apply undifferentiated attention ablation on $h$ (Eq.~\ref{equation:undiff_attention}) \Comment{Mask head influence}
        \State $\mathrm{Imp}(h) \gets \mathbb{E}_{(x, y) \sim \mathcal{D}_{\mathrm{probe}}} [\mathcal{L}(x, y;\mathcal{M}^{-c}_{\mathrm{ref}}) - \mathcal{L}(x, y;\mathcal{M}_{\mathrm{ref}})]$
    \EndFor
    \State $\mathcal{H}_{\mathrm{math}} \gets$ top-$k$ heads by $\mathrm{Imp}(h)$ \Comment{Select reasoning circuits}

    \State $\mathcal{S} \gets \mathbf{0}$ \Comment{Initialize sample scores}
    \For{$(x_i, y_i) \in \mathcal{D}$}
        \For{$h \in \mathcal{H}_{\mathrm{math}}$}
            \State Run $\mathcal{M}_{\mathrm{ref}}$ on $x_i$ (problem only)
            \State Extract $A^h$; normalize rows so $\sum_{j} A^h_{k,j} = 1$
        \EndFor
        \State $\alpha_k \gets \frac{1}{|\mathcal{H}_{\mathrm{math}}|} \sum_{h \in \mathcal{H}_{\mathrm{math}}} \sum_{j=1}^n A^h_{k,j}$
        \State $S(x_i) \gets \frac{1}{n} \sum_{k=1}^n \left(\alpha_k - \frac{1}{n} \sum_{k'=1}^n \alpha_{k'}\right)^2$ \Comment{Variance score}
        \State $\mathcal{S}[i] \gets S(x_i)$
    \EndFor

    \State $p_i \gets \frac{\mathcal{S}[i]}{\sum_{j} \mathcal{S}[j]}$ \Comment{Normalize scores to probabilities}
    \State $m \gets \lfloor \rho \cdot |\mathcal{D}| \rfloor$ \Comment{Subset size}
    \State $\mathcal{D'} \gets$ Soft-sample $m$ examples from $\mathcal{D}$ according to $\{p_i\}$
    \State \textbf{return} $\mathcal{D'}$
\EndProcedure
\end{algorithmic}
\label{alg:circuitseer}
\end{algorithm*}

%% file: chapters/case_study.tex
\tcbset{
    breakable,
    colframe=blue!5!black,
    colback=gray!10!white,
    fonttitle=\bfseries,
    width=\columnwidth 
}

\begin{tcolorbox}[
    title=\textbf{Example of a High-Scoring Sample (High CircuitSeer Score)},
    fonttitle=\bfseries
]
\textbf{Input Problem ($x$):} \\
At the 75th United Nations General Assembly, China pledged to adopt more effective policies and measures to strive to peak carbon dioxide emissions before 2030 and achieve carbon neutrality (referred to as the ``dual carbon goals'') before 2060. This demonstrates China's firm determination to address climate change, indicating that China's economic structure and socio-economic operational models will undergo profound transformations, greatly promoting the clean and green transition of its industrial chain. New energy vehicles, particularly electric vehicles, are key strategic emerging industries that play a crucial role in achieving the dual carbon goals.

To understand the sales of electric vehicles in a certain region, an organization obtained, via the least squares method, the linear regression equation for sales volume $y$ (in units of 10,000 vehicles) with respect to year $x$:
$$
\hat{y} = 4.7x - 9495.2
$$
The sample variance of the sales volume $y$ is $S_y^2 = 50$, and the sample variance of the year $x$ is $S_x^2 = 2$.

\vspace{0.5em}
\textbf{Problem (1)} \\
Calculate the correlation coefficient $r$ between $y$ and $x$, and based on this, determine the strength of the linear relationship between electric vehicle sales volume $y$ and year $x$.

\vspace{0.5em}
\textbf{Problem (2)} \\
The organization also surveyed 100 car owners in the region regarding gender and vehicle type purchased. The data are summarized in the table below:

\begin{center}
\begin{tabularx}{\columnwidth}{@{}l *{3}{>{\centering\arraybackslash}X}@{}}
\toprule
              & Non-electric Cars & Electric Cars & Total \\
\midrule
Male      & 30                & 20            & 50    \\
Female    & 15                & 35            & 50    \\
\midrule
Total     & 45                & 55            & 100   \\
\bottomrule
\end{tabularx}
\end{center}

Can it be concluded, with 99\% confidence, that the purchase of electric cars is associated with gender?

\vspace{0.5em}
\textbf{Problem (3)} \\
Among the owners who purchased electric cars, a stratified sample of 11 individuals was selected according to gender. Then, 2 individuals were randomly selected from these 11. Let $X$ denote the number of males among the 2 selected individuals. Find the probability distribution and the mathematical expectation of $X$.

\vspace{0.5em}
\textbf{Reference Formulas}

\begin{enumerate}
    \item[(i)] Linear regression equation: $\hat{y} = \hat{b}x + a$, where
    $$
    \begin{aligned}
    \hat{b} &= \frac{\sum_{i=1}^{n}(x_i - \overline{x})(y_i - \overline{y})}{\sum_{i=1}^{n}(x_i - \overline{x})^2}, \\
    a &= \overline{y} - \hat{b}\,\overline{x}.
    \end{aligned}
    $$

    \item[(ii)] Correlation coefficient:
    $$
    r = \frac{\sum_{i=1}^{n}(x_i - \overline{x})(y_i - \overline{y})}{\sqrt{\sum_{i=1}^{n}(x_i - \overline{x})^2 \cdot \sum_{i=1}^{n}(y_i - \overline{y})^2}}
    $$
    If $r > 0.9$, it can be judged that $y$ and $x$ are strongly linearly correlated.

    \item[(iii)] Chi-square test statistic for $2 \times 2$ contingency table:
    $$
    \begin{aligned}
    \chi^2 &= \frac{n(ad - bc)^2}{(a+b)(c+d)(a+c)(b+d)} \\
    &\quad \text{where } n = a + b + c + d.
    \end{aligned}
    $$
\end{enumerate}
    
\vspace{1em} 

\textbf{Reasoning Trace (CoT, $r$):} \\
(1) The correlation coefficient $r$ is derived from the regression slope $\hat{b} = 4.7$ and the standard deviations $S_x = \sqrt{2}$, $S_y = \sqrt{50}$. Using the formula $r = \hat{b} \cdot \frac{S_x}{S_y}$, we compute:
$$
r = 4.7 \cdot \frac{\sqrt{2}}{\sqrt{50}} = 4.7 \cdot \frac{1}{5} = 0.94.
$$
Since $r > 0.9$, the linear relationship is strong.
    
(2) For the chi-square test, assign $a=30, b=20, c=15, d=35$. Compute:
$$
\begin{aligned}
\chi^2 &= \frac{100 \cdot (30 \cdot 35 - 20 \cdot 15)^2}{50 \cdot 50 \cdot 45 \cdot 55} \\
       &= \frac{100 \cdot 750^2}{6,\!187,\!500} \\
       &= \frac{56,\!250,\!000}{6,\!187,\!500} \approx 9.09.
\end{aligned}
$$
Since $9.09 > 6.635$ (critical value for $\alpha = 0.01$), we reject independence — gender and car type are associated at 99\% confidence.
    
(3) Among 55 electric car owners (20M, 35F), a proportional stratified sample of 11 yields 4 males and 7 females. Selecting 2 from 11, $X \sim \text{Hypergeometric}(N=11, K=4, n=2)$. Thus:
$$
\begin{aligned}
P(X=0) &= \frac{\binom{7}{2}}{\binom{11}{2}} = \frac{21}{55}, \\
P(X=1) &= \frac{\binom{4}{1}\binom{7}{1}}{\binom{11}{2}} = \frac{28}{55}, \\
P(X=2) &= \frac{\binom{4}{2}}{\binom{11}{2}} = \frac{6}{55}.
\end{aligned}
$$
Expectation: $E[X] = 2 \cdot \frac{4}{11} = \frac{8}{11}$.
    
\vspace{1em}

\textbf{Final Answer ($y$):} \\
(1) Correlation coefficient: $\boxed{0.94}$ — strong linear relationship. \\
(2) Chi-square statistic: $\boxed{9.09}$ — sufficient evidence at 99\% confidence to conclude association between gender and electric car purchase. \\
(3) Probability distribution:
$$
\begin{cases}
P(X=0) = \dfrac{21}{55}, \\
P(X=1) = \dfrac{28}{55}, \\
P(X=2) = \dfrac{6}{55},
\end{cases}
\quad \text{Expectation: } \boxed{\dfrac{8}{11}}.
$$
\end{tcolorbox}
\vspace{1em} 

\begin{tcolorbox}[
    title=\textbf{Example: Integer Values of a Power Sum Expression (High CircuitSeer Score)},
    fonttitle=\bfseries
]
\textbf{Input Problem ($x$):} \\
How many positive integers $n$ are there for which $\left(2022 + \frac{1}{2}\right)^n + \left(25 + \frac{1}{2}\right)^n$ is an integer?

\vspace{0.5em}

\textbf{Reasoning Trace (CoT, $r$):}

First, express the numbers as fractions:
$$
2022.5 = \frac{4045}{2},\quad 25.5 = \frac{51}{2}
$$
So the expression becomes:
$$
\left(\frac{4045}{2}\right)^n + \left(\frac{51}{2}\right)^n = \frac{4045^n + 51^n}{2^n}
$$
We require $2^n$ divides $4045^n + 51^n$.

Both $4045$ and $51$ are odd, so their powers are odd, and their sum is even (divisible by $2$). But for higher $n$, we need divisibility by higher powers of $2$.

Using the Lifting The Exponent Lemma (LTE) for $p=2$:
Since $4045 \equiv 1 \pmod{4}$ and $51 \equiv 3 \pmod{4}$, $4045 \equiv -51 \pmod{4}$.
For odd $n$, LTE gives $v_2(4045^n + 51^n) = v_2(4045 + 51) + v_2(n) = 12 + v_2(n)$,
since $4045 + 51 = 4096 = 2^{12}$.

For $2^n$ to divide $4045^n + 51^n$, we need $12 + v_2(n) \geq n$.
\begin{itemize}
    \item For $n = 1$, $v_2(1) = 0$, $12 \geq 1$: true.
    \item For $n = 3,5,7,9,11$, $v_2(n)=0$, so $12 \geq n$.
\end{itemize}
So, all odd $n \leq 11$ are valid.

For even $n$, the sum modulo $4$ is $2$, not divisible by $4$,
so for $n \geq 2$ even, the sum is not divisible by $2^n$.

Thus, the valid $n$ are $1,3,5,7,9,11$.

\vspace{0.5em}

\textbf{Final Answer ($y$):}

There are $\boxed{6}$ positive integers $n$ for which $\left(2022 + \frac{1}{2}\right)^n + \left(25 + \frac{1}{2}\right)^n$ is an integer.
\end{tcolorbox}
\vspace{1em} 

\begin{tcolorbox}[
    title=\textbf{Example: Only Even Prime Number (Low CircuitSeer Score)},
    fonttitle=\bfseries
]
\textbf{Input Problem ($x$):} \\
Find all even primes.

\vspace{0.5em}

\textbf{Reasoning Trace (CoT, $r$):}

Prime numbers are defined as natural numbers greater than 1 with no positive divisors other than 1 and themselves. The only even prime number is 2, because:
\begin{itemize}
    \item All even numbers (except 2) are divisible by 2, so they have at least three divisors: 1, 2, and themselves.
    \item 2 has exactly two divisors: 1 and 2.
\end{itemize}
For any even number $p > 2$, $p = 2k$ for $k > 1$, so $p$ is divisible by 1, 2, and $p$, hence not prime.

Thus, 2 is the only even prime.

\vspace{0.5em}

\textbf{Final Answer ($y$):}

The only even prime is $\boxed{2}$.
\end{tcolorbox}
\vspace{1em} 

\begin{tcolorbox}[
    title=\textbf{Example: Reciprocal of \texorpdfstring{$-1$}{-1} (Low CircuitSeer Score)},
    fonttitle=\bfseries
]
\textbf{Input Problem ($x$):} \\
The reciprocal of $-1$ is\underline{\hspace{3cm}}.

\vspace{0.5em}

\textbf{Reasoning Trace (CoT, $r$):}

The reciprocal of a number $x$ is defined as the number $y$ such that $x \cdot y = 1$. For $x = -1$:
$$
(-1) \cdot y = 1 \implies y = \frac{1}{-1} = -1.
$$
Verification: $(-1) \times (-1) = 1$, which satisfies the definition.

Thus, the reciprocal of $-1$ is $-1$.

\vspace{0.5em}

\textbf{Final Answer ($y$):}

The reciprocal of $-1$ is $\boxed{-1}$.
\end{tcolorbox}